\documentclass{article}
\usepackage[preprint,main]{iclr2027_preprint}
\usepackage{times}
\usepackage[T1]{fontenc}
\usepackage{amsmath,amssymb}
\usepackage{booktabs}
\usepackage{graphicx}
\usepackage{array}
\usepackage{hyperref}
\usepackage{url}
\usepackage{tabularx}

\title{State Representation Shapes Long-Horizon Agent Learning and Transfer: An Empirical Study}
\author{  Dong Nie \\
  Independent Researcher \And
  Rui Lan \\
  Independent Researcher \And
  Yixin Nie \\
 Independent Researcher \And
  Lingzi Hong \\
  University of North Texas
  }

\newcommand{\Full}{\textsc{Full}}
\newcommand{\Compact}{\textsc{Compact}}

\begin{document}
\maketitle

\begin{abstract}
An end-to-end success rate alone cannot distinguish failure to learn an action
from failure to apply it as task state changes. We study this distinction in deterministic
tool workflows that provide the current artifact state and candidate action
recipes. Using 1.7B, 8B, and 27B Qwen models, we compare supervised learning
across workflow horizons and cross the state representation used for training
with the one used at evaluation. Horizon-matched supervision yields high
completion, but an H4-trained 1.7B model transfers poorly to longer workflows
under full-state observations. In an H16 crossover, compact-trained 1.7B
adapters complete 50/50 held-out tasks in compact view and 0/50 in full view
for each of three paired training seeds; full-trained outcomes vary across
seeds. A separate 8B adapter pair shows a similar asymmetry. The view
dependence also appears in paired-choice tasks where both candidate calls are
executable and the correct choice depends on table contents; here, the compact
mask does not consult the target action. An explicit relevance cue improves
full-view execution but does not close the gap for the tested checkpoint.
In a separate 27B ScienceWorld study, compact-state training improves local
accuracy under both views, indicating that the transfer pattern depends on the
task and observation construction. Together, these experiments identify representation-sensitive transfer
as a bottleneck in the controlled workflows, without isolating content
reduction from other changes to the observation.
\end{abstract}

\section{Introduction}
\label{sec:intro}

\emph{What limits long-horizon agent learning?} Successful long-horizon behavior depends on persistent goal and state tracking, state-dependent action selection, valid tool execution, and recovery when errors occur. We characterize three of these demands through a controlled empirical study: selecting actions given the current state, issuing valid tool calls, and recovering after rejected calls. Our multi-step workflows explicitly provide the current artifact state and candidate action recipes with specified arguments, allowing us to focus on the use of available state rather than memory-based state reconstruction. By varying supervised training and state representation, we examine how these behaviors change with learning, whether improvements transfer across horizons and state views, and how errors and recovery relate to task completion.

We focus on state representation because longer workflows change not only the number of required actions but also the information available when each action is selected. As tools produce intermediate artifacts, the agent must identify which available inputs and recipes are relevant to the next step. This raises a learning question: does supervised training produce behavior that transfers across horizons and state views, or behavior that depends on the particular representation encountered during training? We evaluate transfer from shorter to longer workflows and separately cross training and evaluation representations. In the representation crossover, the training examples share the same underlying task instances and target actions, and the resulting policies are evaluated on common held-out tasks under both state views. Gold-action supervision allows us to examine these effects under explicit action-level guidance; it does not independently test exploration or temporal credit assignment.

Our setting delegates table computation to deterministic tools and verifies intermediate and final artifacts exactly. Given the supplied action recipes, the agent must select an appropriate recipe for the current state, generate the corresponding tool call with its specified arguments, and complete the workflow within a fixed action budget. We use four-, eight-, and sixteen-action workflows to examine learning, transfer, and execution. Training curves track performance across optimization steps, while evaluation beyond the training horizon and the train/test representation crossover assess transfer. Correct-prefix probes measure action selection from correct intermediate states, while free-running rollouts measure completion, errors, and recovery under the policy's own decisions. Comparisons across untuned and fine-tuned checkpoints track changes in local action-selection accuracy and end-to-end performance under each training condition.

The results separate learning at the training horizon from transfer to a new
horizon or state view. Horizon-matched supervision supports high H16
completion at all three model sizes, but the H4-trained 1.7B adapter fails on
longer full-state workflows. More decisively, the H16 representation crossover
yields the same compact-to-full failure at the endpoint across three paired
1.7B training seeds, despite variation in the full-trained results. One
held-out 8B adapter pair shows a similar asymmetry. The original
compact-trained learning curve's decline in full-view accuracy, however, does
not recur uniformly across seeds; the replicated result concerns endpoint
transfer, not that particular learning trajectory.

Several controls delimit this finding. Matching an explicit relevance cue
across full and compact views improves full-view execution without closing the
gap for the tested H4-trained adapter. Training with a recency-based mask that
does not use the reference action also yields poor transfer to full state. In
a separate task family, two calls are executable and a changed table value
determines which advances the goal. Compact-trained models solve both variants
of every held-out pair in compact view, yet generalize poorly to full view;
full-view training remains near chance under the tested budget. Correct-prefix
and free-run evaluations further show that action agreement, completion, and
recovery need separate measurement. Our ScienceWorld study gives a different
cross-view pattern, indicating that the artifact-workflow result should not be
assumed to hold across observation constructions. The main compact view
retains inputs of the evaluator-selected next action, so its benefit cannot be
assigned solely to hiding other artifact contents.

\section{Related Work}
\label{sec:related}

\paragraph{Agent learning and horizon transfer.}
AgentTuning improves agent capabilities through supervised
training on interaction trajectories mixed with general
instruction data \citep{zeng2024agenttuning}.
A central challenge in sequential imitation learning is the
mismatch between expert-induced states and those encountered
under a learned policy \citep{ross2011reduction}.
On-policy expert corrections address this mismatch in
multi-turn language-model agents by collecting expert
continuations after student interactions
\citep{lauffer2025imitation}.
\citet{kim2026training} study reinforcement-learning stability
and horizon generalization, investigating macro-actions and
subgoal decomposition to reduce the effective horizon.
Our study examines supervised learning with supplied artifact
states and action recipes, evaluating transfer across both
workflow horizons and state representations.

\paragraph{Tool interfaces and observation relevance.}
SWE-agent demonstrates how agent--computer interface design
affects software-engineering performance
\citep{yang2024swe}, while CodeAct studies executable
Python as an action representation \citep{wang2024executable}.
Controlled comparisons of coding-agent tool architectures
further examine effects on consistency, exploration, and
efficiency \citep{xu2026devil}.
Observation content also affects performance.
Lost in the Middle documents sensitivity to the position of
relevant evidence \citep{liu2024lost}.
GSM-DC examines reasoning under systematically introduced
irrelevant information and the effects of training with
distractors \citep{yang2025llm}, while MUDDLE compares
topically similar distractors with length-matched random
documents \citep{luo2026muddle}.
Our intervention changes the visible artifact contents while
keeping the task recipes and action targets fixed.

\paragraph{Observation management and representation transfer.}
HiAgent organizes working memory around subgoals
\citep{hu2025hiagent}.
Within SWE-agent, observation masking can achieve solve
rates competitive with language-model summarization
\citep{lindenbauer2025complexity}.
Learning-based approaches include Memory-as-Action, which
trains context-editing actions, and CompactionRL, which
jointly optimizes task execution and summary generation
\citep{zhang2026memory,li2026compactionrl}.
TRACE evaluates compression events through paired
continuations from the same environment state and optimizes
a compression prompt while keeping the agent's model
parameters fixed \citep{min2026toward}.
We train the agent on matched examples under different state
views, then evaluate each checkpoint under both views on
common held-out tasks. This crossover tests whether
supervised gains transfer across representations.
Because our compact view retains the contents of inputs selected
by the evaluator, it supplies relevance information and serves as
an assisted diagnostic, not an autonomous memory policy.

\paragraph{Execution reliability and recovery.}
$\tau$-bench evaluates goal-state correctness and consistency
across repeated trials \citep{yao2024taobench}.
Broader reliability analyses distinguish consistency,
robustness, predictability, and safety from aggregate task
accuracy \citep{rabanser2026towards}.
Recovery benchmarks introduce controlled environmental
failures: Hell or High Water makes functions unavailable
while preserving alternative solutions
\citep{wang2025hell}, and ToolMaze evaluates replanning under
controlled tool perturbations \citep{zhu2026tools}.
We measure completion after agent-generated calls are rejected
in otherwise deterministic workflows, for both untuned and
fine-tuned checkpoints. Policies can encounter different
errors with different remaining budgets, so these rates
describe observed recovery without comparing policies under
matched failures.

\section{Experimental Design}
\label{sec:design}

We compare learning at the training horizon, transfer to longer
workflows, and transfer between state views. Correct-prefix
probes and autonomous rollouts measure local decisions and
task completion; supplied-recipe probes assess tool-call
generation when the required action is given.
Appendix~\ref{app:implementation} gives the detailed protocol.

\paragraph{Datasets.}
The controlled experiments use our procedurally generated
table-workflow dataset. Each supervised example pairs an
artifact-state observation and candidate recipes with the
gold next action; held-out task instances support evaluation.
External evaluation uses sampled ScienceWorld trajectories
and task variations \citep{wang2022scienceworld}.
We will publicly release the generated workflow dataset,
including task definitions, supervised examples, and
evaluation split specifications.

\subsection{Deterministic artifact workflows}
\label{sec:workflows}

A task is a directed acyclic graph of $H$ required operations
with artifact dependencies. The main family uses chains
with $H\in\{4,8,16\}$; a second family uses shallow branching.
Horizon counts required operations, not dependency depth or
attempted calls. Deterministic tools map numeric columns,
sort or filter rows, and select columns. In the main horizon
experiments, the agent receives the accumulated artifact
state, remaining budget, and two candidate recipes: the
evaluator-selected next recipe and one distractor from the
same task. Recipes specify the tool, input/output
identifiers, and parameters; the agent selects a recipe
and returns its JSON call. Table computation is delegated
to tools.

Candidate placement and distractor selection are deterministic
functions of task seed and stage. The target is not explicitly
labeled, although including it among two candidates provides
assistance. On chains, dependencies and artifact availability
identify the unique executable unfinished operation.

Success requires exact verification of all required
artifacts and their dependency ancestry. Malformed calls,
recipe mismatches, repeated outputs, and calls with missing
prerequisites are rejected. Rejected calls consume budget,
leave artifacts unchanged, and expose an error in the next prompt. Autonomous artifact evaluations allow
$2H$ calls, including rejections. Recovery therefore
concerns retrying after rejection, not repairing corrupted
artifacts. On branching tasks, a valid ready action can
differ from the canonical target; canonical disagreement
is not necessarily an execution error.

\paragraph{Content-dependent action choice.}
A separate H16 family gives the agent a fixed goal and two
executable recipes at each stage: remove negative-valued rows
from the active table or sort it by key. Both recipes use the
same input and output identifiers. The goal requires filtering if
the active table contains a negative value and sorting otherwise.
For each paired task, changing one cell between $-1$ and $+1$ at one
stage flips the correct recipe while preserving the goal,
candidate order, names, and earlier states. In this family,
compact observations retain the active input's contents and
all artifact names while hiding other contents. Retention
does not consult the correct action, since both recipes share
the active input. A wrong executable call advances the stage
but prevents verified success. Stage inputs are independent;
this control tests repeated content-dependent choices, not
dependency composition.

\subsection{Full and compact observations}
\label{sec:views}

Let $A_t$ map available artifact names to their contents.
The \Full{} view displays the contents of all available
artifacts. It represents the current accumulated artifact
state, rather than the preceding response transcript.
Prompts are reconstructed at each interaction, including
the previous tool error when present.
\Compact{} displays artifact contents only for the reference
action's inputs $I_t^*$:
\begin{equation}
A_t^{\mathrm{compact}}(k)=
\begin{cases}
A_t(k), & k\in I_t^*,\\
\texttt{available}, & k\notin I_t^*.
\end{cases}
\label{eq:compact}
\end{equation}
The reference is the gold action during SFT and the
evaluator-selected next action during evaluation. Both
views retain all artifact names; \texttt{available} marks
an existing artifact whose contents are hidden. For the same
state, the recipes, transition rules, and verifier are
unchanged. On branching tasks, retention follows one canonical ready
node. This intervention both reduces the amount of displayed
artifact content and supplies relevance information. Attributing
gains specifically to reduced distraction from displayed
artifact content would require a relevance-matched control.
The separate paired-choice control retains the
explicit active input rather than using $I_t^*$ in
Equation~\ref{eq:compact}.

\paragraph{Relevance-control views.}
For fixed checkpoints, \Full{} and \Compact{} are also evaluated
with the same explicit \texttt{relevance\_hint}, which lists the
input artifact names of the evaluator-selected next action.
This matches the explicit cue across views while changing which
artifact contents are visible. A \texttt{recent2} view instead
retains the contents of the two most recently created available
artifacts, based on successful creation history and without
consulting the reference action. All control views preserve the
same candidate recipes; that pair still includes the evaluator-selected
action. The explicit-cue controls are inference-time diagnostics of fixed
checkpoints. \texttt{recent2} is also used as an H16 training view
in a separate crossover. Its content mask is reference-free,
but the candidate pair still includes the evaluator-selected
action. The H4-trained reference adapter and the original H16
adapters are evaluated on the same first 20 tasks in the frozen
H16 test set for the inference-time controls.

\paragraph{Illustrative state.}
Suppose an older artifact \texttt{s0} and an intermediate
result \texttt{a1} are available, and the next action maps
\texttt{a1} into the absent output \texttt{a2}. Full state
shows both tables. Compact state retains the contents of
\texttt{a1} and replaces those of \texttt{s0} with
\texttt{available}, preserving both names and the same
recipes. If a distractor uses \texttt{s0}, this visibility
pattern can help identify the relevant input without an
explicit target label. Both views still invoke tools on
the same underlying tables; masking does not remove an
artifact from the environment.

\subsection{Learning and evaluation protocol}
\label{sec:protocol}

We evaluate untuned checkpoints and horizon-specific
adapters of Qwen3-1.7B, Qwen3-8B \citep{yang2025qwen3},
and Qwen3.8-27B. Untuned denotes no workflow-specific SFT.
Artifact SFT pairs gold-prefix observations with JSON
actions and uses completion-only loss, 100 optimizer steps,
learning rate $2\times10^{-4}$, and adapter rank 16.
Each update presents eight examples, except the 27B H16
run, which presents 64. The 1.7B adapters use
LoRA \citep{hu2021lora}; 8B and 27B use NF4 QLoRA
\citep{dettmers2023qlora}. The 1.7B and 8B horizon-training datasets each
contain 800 action examples: 200, 100, and 50 complete
workflows at H4, H8, and H16, respectively. Thus equal action
counts do not match workflow counts, token exposure, or
compute. The 27B H4/H8/H16 runs retain 800/1,600/3,200
source rows and present 800/800/6,400 examples, respectively
(Appendix~\ref{app:implementation}). Differences in model
generation, adaptation, data exposure, and inference paths
preclude isolating parameter-count effects.

Learning curves evaluate fixed held-out tasks at the training horizon. To test
horizon transfer, we hold the adapter and observation view fixed while
increasing task length; separately trained horizon-matched adapters answer a
different question. The H16 representation crossover trains full- and
compact-view 1.7B adapters on the same 800 gold actions from 50 workflows.
Three paired training seeds are evaluated on 50 common held-out H16 workflows
under both views. We also test one full/compact 8B adapter pair on these
workflows and train three additional 1.7B adapters with the target-independent
\texttt{recent2} view. The matched-cue and state-load controls evaluate fixed
checkpoints (Appendices~\ref{app:statistics} and~\ref{app:horizon-state}).

For content-dependent choice, we train three paired 1.7B full/compact adapter
pairs on 100 training task pairs (200 workflows) and evaluate 50 held-out pairs. We rank two
equal-length candidate calls by conditional likelihood at the pivotal state;
free-running rollouts use the first ten held-out pairs. This forced choice
differs from generated-call accuracy and autonomous completion. Training data,
seeds, budgets, and probe construction are detailed in
Appendix~\ref{app:content-choice}.

\emph{Oracle-prefix accuracy} measures canonical action
agreement in separately restored gold-prefix states;
\emph{clean oracle-path rate} requires agreement at every
probed step. A mistaken prediction does not affect the next
probe, which again starts from its gold prefix; clean
oracle-path rate is therefore not an autonomous success
rate.

\emph{Free-run success} instead requires
verified completion under the policy's own calls.
\emph{Final progress} is the fraction of required nodes
verified, regardless of execution order. We also measure
mean attempted calls, episode-level rejection incidence,
and \emph{observed recovery}: eventual success conditional
on at least one rejected call. Recovery is undefined when
no episode contains a rejection. Mean call counts include
all attempts and are averaged over all episodes. For a controlled post-error comparison, we evaluated the seed-20260920 Qwen3-8B H16 full/compact adapter pair on 50 common workflows in each train-view by test-view condition. Each episode started after four verified gold actions; the evaluator then issued the same rejected call, which consumed one call without changing artifacts. The policy selected subsequent calls autonomously with 27 of the 32-call budget remaining. The 200 episodes cover the four train/test-view combinations. No-injection baseline episodes follow model-generated states and are descriptive, not a paired estimate of the injected error's effect.

Artifact evaluation uses greedy decoding with thinking disabled and
at most 160 generated tokens per call. Shared learning
curves evaluate steps 20, 40, 60, 80, and 100 on ten tasks
for 1.7B/8B and twenty for 27B; additional 27B H16 probes
at steps 5 and 10 assess early learning.
Original endpoint comparisons use 20 tasks per condition;
within-run view contrasts pair the same tasks. The training
pilot uses ten development tasks and one training seed;
ScienceWorld uses three training seeds, natural reference-path
lengths, and environment-specific observation views. Its
100-call budget, score-100 success criterion, and parser-rejection
measure differ from the artifact protocol. We do not assume
its views implement Equation~\ref{eq:compact}; protocol and
split details appear in Appendix~\ref{app:scienceworld}.

Original-crossover paired analyses use tasks as sampling units
and condition on the trained adapters. For the expanded H16
crossover, we report the paired train-view by test-view interaction
separately for each of the three seed pairs, along with their
descriptive mean and sample standard deviation; seed-by-task
outcomes are not pooled as independent observations. Matched-cue
comparisons preserve task pairing, but their exact tests are
exploratory and unadjusted across the control comparisons.
Untuned and tuned 27B runs use different tasks
and inference paths, so their comparison is descriptive.
Appendices~\ref{app:implementation} and~\ref{app:statistics}
provide metric definitions, checkpoint schedules, and
statistical procedures.

\section{Results}
\label{sec:results}

\subsection{Learning and horizon transfer}
\label{sec:learnability}
\label{sec:27b-learning}

\textbf{Horizon-matched supervision yields high completion across
all three model sizes} (Table~\ref{tab:learnability}). Separate
27B adapters achieve 20/20 completions at each of H4, H8,
and H16 under both views, using exactly $H$ calls without
rejections (Appendix~\ref{app:27b-learning}). Thus, the evaluated workflows can be completed without
rejections under full observations after horizon-matched
training.

At H16, oracle accuracy for 1.7B rises from 71.9\% at step
20 to 99.4\% at step 100; 8B reaches 99.4\% by step 20.
The 27B H16 probes reach 320/320 correct actions by step 10
(640 action-example presentations). All three 27B adapters are
at ceiling on the common step-20--100 grid
(Appendix~\ref{app:controls}). However, 27B H16 uses 64
examples per update versus eight in the smaller-model runs.
The unequal exposure and differences in model configuration
prevent attributing early saturation to model size alone
or ranking sample efficiency from these curves.

The untuned 1.7B baseline already agrees with individual
oracle actions on 42.5\%, 36.9\%, and 35.9\% of H4, H8,
and H16 probes, yet completes only 2/20, 0/20, and 0/20
tasks.

\textbf{Learning short workflows does not ensure transfer}. The 1.7B
model reaches perfect H4 performance across five training
seeds. The adapter used for horizon transfer completes only
5/20 H8 and 0/20 H16 tasks under full observations. Its
post-stage-four oracle accuracy also declines, despite
restoring correct prefixes before each probe. The 27B rows
evaluate each adapter at its training horizon, so they do
not establish whether H4- or H8-trained 27B behavior
transfers to H16.

\begin{table}[t]
\centering
\caption{Full-state free-run success on 20 tasks per cell. Each horizon-matched row reports one adapter per horizon; the three-seed H16 crossover appears in Table~\ref{tab:crossover-seeds}. The 27B rows use step 100. Only the H4-trained row measures short-to-long transfer. Model configurations differ across rows.}
\label{tab:learnability}
\begin{tabular}{lccc}
\toprule
Checkpoint and training & H4 & H8 & H16\\
\midrule
1.7B, untuned & 2/20 & 0/20 & 0/20\\
1.7B, H4 SFT & 20/20 & 5/20 & 0/20\\
1.7B, horizon-matched SFT & 20/20 & 20/20 & 17/20\\
8B, horizon-matched SFT & 20/20 & 20/20 & 20/20\\
27B, horizon-matched SFT & 20/20 & 20/20 & 20/20\\
\bottomrule
\end{tabular}
\end{table}

\subsection{Test-time state representation}
\label{sec:rescue}

\textbf{Changing only the observation view substantially improves
chain execution for fixed checkpoints} (Table~\ref{tab:rescue}).
For the H4-trained 1.7B adapter, compact observations raise
H16 success from 0/20 to 18/20, with the same contrast on
twenty fresh tasks. This replicates across tasks, not
training runs. Untuned 8B likewise improves from 0/20 to
16/20 at H16. Untuned 27B shows the same direction of improvement as 8B:
correct-prefix accuracy increases, and observed free-run
success rises from 19/20 to 20/20 at H8 and from 16/20 to
20/20 at H16 (Appendix~\ref{app:statistics}).
Compaction also supplies relevance information
(Equation~\ref{eq:compact}), so these gains cannot be
attributed to hiding artifact contents alone.

The observation change also reduces execution cost for
untuned 8B: mean attempted calls fall from 15.75 to 10.30
at H8 and from 32.00 to 20.35 at H16, with the checkpoint,
required operations, and call budget fixed
(Appendix~\ref{app:chaincontrasts}).

\begin{table}[t]
\centering
\caption{Paired full--compact evaluation of fixed checkpoints on 20 tasks per cell. Post-4 oracle accuracy covers stages 5--$H$. Training status, model generation, and inference paths differ across rows; these are observation contrasts, not isolated model-size effects.}
\label{tab:rescue}
\begin{tabular}{llcc}
\toprule
Checkpoint & Horizon & Post-4 oracle (\%) & Free-run success\\
 & & Full $\to$ compact & Full $\to$ compact\\
\midrule
1.7B, H4 SFT & 8 & 68.8 $\to$ 100.0 & 5/20 $\to$ 20/20\\
1.7B, H4 SFT & 16 & 50.4 $\to$ 98.3 & 0/20 $\to$ 18/20\\
8B, untuned & 8 & 58.8 $\to$ 96.3 & 1/20 $\to$ 15/20\\
8B, untuned & 16 & 55.8 $\to$ 91.3 & 0/20 $\to$ 16/20\\
27B, untuned & 8 & 92.5 $\to$ 100.0 & 19/20 $\to$ 20/20\\
27B, untuned & 16 & 82.1 $\to$ 99.2 & 16/20 $\to$ 20/20\\
\bottomrule
\end{tabular}
\end{table}

A fixed-checkpoint control shows that greater
displayed state alone does not reproduce the horizon decline.
Adding eight irrelevant 32-row tables at H8 raises mean
correct-prefix input length from 542 to 2,390 tokens, yet
action agreement changes from 138/160 to 136/160 and free-run
success from 8/20 to 7/20; other loads are nonmonotone. On
prefix-matched H16 tasks, Recent-2 improves correct-prefix
agreement from 230/320 to 294/320 (19/20 paired seeds improve),
but free-run success reaches only 4/20 versus 0/20 under full
state. Under Recent-2, H4 success is 19/20. Limiting contents
improves local decisions without eliminating long-horizon
failure. Recency can cue relevance and names still accumulate,
so this does not isolate a pure decision-count effect
(Appendix~\ref{app:horizon-state}).

\subsection{Representation-dependent learning}
\label{sec:crossover}

\begin{figure}[t]
\centering
\includegraphics[width=1\linewidth]{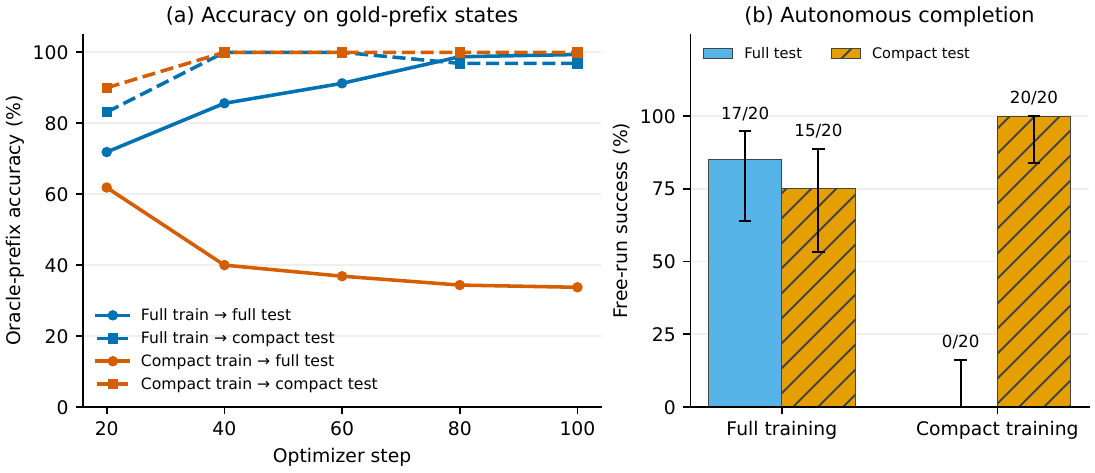}
\caption{Original-seed H16 representation crossover. Left: oracle accuracy on ten paired tasks per checkpoint. Right: completion on twenty tasks. The full-view accuracy decline shown is specific to this training pair; its magnitude and direction varied across the two added seeds. Wilson 95\% intervals describe task uncertainty conditional on the adapters.}
\label{fig:crossover}
\end{figure}

The original 1.7B H16 crossover showed a large dependence of
completion on the training and evaluation views
(Figure~\ref{fig:crossover}). We repeated the endpoint comparison
with two additional training-seed pairs, evaluating all three
pairs on 50 common held-out workflows. Each compact-trained
adapter completed 50/50 tasks under compact observations and
0/50 under full observations. Full-trained completion varied
across seeds: 35--49/50 under full and 24--50/50 under compact
observations (Table~\ref{tab:crossover-seeds}). 
The train-view by test-view completion interaction was 142,
98, and 78 percentage points for seeds 20260910, 20260926,
and 20260927, respectively (mean 106 points; sample SD 32.7).
These are descriptive seed-level estimates from three training
pairs, not a population-level test. The full-trained outcomes
show why the interaction should not be read as a universal
ranking of training views.

\begin{table}[t]
\centering
\small
\caption{H16 training-representation crossover on the same 50 held-out test workflows. Each cell is verified free-run completion. The first seed is the original historical pair; the other two pairs completed training uninterrupted. $\Delta$ is the paired train-view by test-view difference of differences, in percentage points.}
\label{tab:crossover-seeds}
\begin{tabular}{lrrrrr}
\toprule
Seed & $F\!\to\!F$ & $F\!\to\!C$ & $C\!\to\!F$ & $C\!\to\!C$ & $\Delta$\\
\midrule
20260910 & 45/50 & 24/50 & 0/50 & 50/50 & 142\\
20260926 & 49/50 & 50/50 & 0/50 & 50/50 & 98\\
20260927 & 35/50 & 46/50 & 0/50 & 50/50 & 78\\
\bottomrule
\end{tabular}
\end{table}

The original ten-task learning curve showed compact-view oracle
accuracy rising from 90.0\% at step 20 to 100\% at step 40,
while full-view accuracy fell from 61.9\% to 33.8\% at step
100. That full-view change was not consistent across the three
seed pairs: step-20-to-100 changes were $-29.4$, $-3.1$, and
$+11.9$ percentage points. The endpoint crossover replicated;
the specific decline in the original learning curve did not
recur uniformly. Figure~\ref{fig:crossover} therefore reports
the original learning curve, while the new endpoint evaluations
use 50 common test workflows.

A single-seed Qwen3-8B H16 crossover shows a similar,
asymmetric endpoint pattern (Table~\ref{tab:crossover-8b}).
The full-trained adapter completed all 50 workflows under both
views, with 800/800 canonical actions correct and no rejected
calls. The compact-trained adapter completed 49/50 workflows
under compact observations (799/800 actions correct), but 0/50
under full observations (336/800 actions correct); all 50
full-view episodes incurred a rejected call and reached the
32-call budget. Thus, compact-trained behavior transferred poorly
to full observations in this adapter pair, whereas full-trained
behavior transferred to compact observations. This is an
additional task-set evaluation, not a training-seed replication;
the compact view retains evaluator-selected inputs, so the
contrast does not isolate artifact-content masking alone.

\begin{table}[t]
\centering
\small
\caption{Qwen3-8B H16 training-view crossover on 50 common workflows (one adapter pair, training seed 20260920). Oracle accuracy is out of 800 decisions; all other counts are out of 50 workflows. Calls are mean attempted calls; rejects count episodes with at least one rejected call.}
\label{tab:crossover-8b}
\begin{tabular}{lcccc}
\toprule
Train $\to$ test & Oracle & Clean & Success & Calls / rejects\\
\midrule
Full $\to$ full & 800/800 & 50/50 & 50/50 & 16.00 / 0\\
Full $\to$ compact & 800/800 & 50/50 & 50/50 & 16.00 / 0\\
Compact $\to$ full & 336/800 & 0/50 & 0/50 & 32.00 / 50\\
Compact $\to$ compact & 799/800 & 49/50 & 49/50 & 16.32 / 1\\
\bottomrule
\end{tabular}
\end{table}

A matched-cue control addresses whether the compact view's
relevance information alone accounts for its effect. On 20 paired
H16 test workflows, an H4-trained adapter completed 0/20 in full
view and 7/20 in full view with an explicit input-name cue; under
the same cue it completed 19/20 in compact view. A target-independent
view retaining the two most recently created artifacts completed
0/20, like the uncued full view. The cue partly improved full-view
execution, but did not close the observed view gap for this
checkpoint. These fixed-checkpoint controls remain assisted because
the displayed candidate pair includes the evaluator-selected action;
they do not establish a general compaction method. Full control
results and the original H16 adapter controls appear in
Appendix~\ref{app:statistics}.

We then train with the reference-free \texttt{recent2}
mask. Across three H16 training seeds, the new adapters
completed 38/50, 44/50, and 43/50 held-out workflows under
\texttt{recent2}, but only 2/50 under full state for every
seed. All three completed 50/50 under the reference-selected
compact view. These comparisons use the same source actions
and test workflows as the full/compact crossover. The full-view transfer failure therefore persists when the
content mask does not consult the reference action. Recency
may still identify useful inputs in a chain, and the candidate
pair remains assisted. This control does not isolate content
reduction from implicit relevance cues or establish a general
compaction policy.

\subsection{\texorpdfstring{Content-dependent selection}{Content-dependent selection}}
\label{sec:content-choice}

The paired-choice control tests whether view dependence
persists when availability cannot identify the right recipe.
At the pivotal stage, both recipes execute, and one changed
table value reverses which meets the goal. Across all three
1.7B training seeds, compact-trained adapters choose correctly
for both variants of every held-out pair under compact
observations (50/50 pairs) and complete 20/20 greedy free
runs per seed (Table~\ref{tab:content-choice}). In full view,
the same adapters choose correctly for both variants of only
9/50 pairs and complete 0/20 free runs. Full-trained adapters
score 49--52/100 pivotal choices under their matching full
view, near the untuned model's 50/100, and complete 0/20
free runs. Thus compact training supports the content-dependent
choice under this budget, but its learned behavior does not
transfer reliably to the full view. The full-trained adapters'
low matched-view accuracy prevents a symmetric transfer claim.

\begin{table}[t]
\centering
\small
\caption{H16 paired-choice control with three Qwen3-1.7B
training seeds. Entries give per-seed counts or their range,
not pooled independent trials. Pivotal choice ranks two
supplied, equal-length calls on 50 pairs (100 decisions);
``Both'' requires both variants correct. Greedy free runs use
ten pairs (20 tasks) per adapter/view.}
\label{tab:content-choice}
\begin{tabular}{lccc}
\toprule
Train $\to$ test & Pivotal /100 & Both /50 & Free run /20\\
\midrule
Full $\to$ full & 49--52 & 0--3 & 0\\
Full $\to$ compact & 53--61 & 3--12 & 0\\
Compact $\to$ full & 59 & 9 & 0\\
Compact $\to$ compact & 100 & 50 & 20\\
\bottomrule
\end{tabular}
\end{table}

Unlike the main chain tasks, this compact mask is
independent of the correct recipe. The control therefore shows
a view-dependent failure on choices that require table contents,
without the original reference-action retention cue. It does
not identify whether the full-view difficulty comes from
additional irrelevant contents, their position, or prompt
length. Its independent stage inputs also limit claims about
dependency composition.

\subsection{Mixed-view training as a transfer control}
\label{sec:augmentation}

A follow-up 1.7B pilot tests whether view mixing mitigates
this asymmetry. Four adapters use the same 800 H16 action
examples, one new training seed, and ten held-out
development tasks. Mixtures assign 400 examples to each
view; random masking does not consult the target action
(Appendix~\ref{app:augmentation}). The additional block-masked
evaluation hides the contents of a contiguous block of half
the available artifacts, ordered by name, without consulting
the target. It preserves
artifact names and recipes while testing a masking rule
not used during training.

\begin{table}[t]
\centering
\small
\setlength{\tabcolsep}{4pt}
\caption{Qwen3-1.7B H16 pilot at step 100. Cells report canonical oracle accuracy (\%); free-run successes/10. Each cell uses 160 probes and ten common development tasks with one training seed. Block masking is target-independent; compact masking is target-dependent.}
\label{tab:augmentation}
\begin{tabular}{lccc}
\toprule
Training condition & Full & Compact & Block-masked\\
\midrule
Full only & 98.75; 8/10 & 95.00; 6/10 & 97.50; 7/10\\
Compact only & 70.00; 0/10 & 100.00; 10/10 & 70.63; 0/10\\
Full/compact mixture & 81.88; 1/10 & 100.00; 10/10 & 76.88; 1/10\\
Full/random mixture & 97.50; 6/10 & 98.13; 7/10 & 96.25; 5/10\\
\bottomrule
\end{tabular}
\end{table}

The full/compact mixture retains 10/10 compact completions
but reaches only 1/10 under full observations
(Table~\ref{tab:augmentation}). Random masking yields higher
observed full- and block-view completion than this mixture,
but does not consistently beat full-only training. None
of the six paired success contrasts against these two
baselines is conclusive. Simple mixing therefore does not
resolve transfer under this budget, and the pilot does not
establish an advantage for random masking. Both mixtures
halve full-view exposure; all tasks have H16. Consequently,
the study neither isolates relevance-cue dependence nor
tests improved short-to-long transfer.

\subsection{Local accuracy, completion, and recovery}
\label{sec:local}
\label{sec:branching}
\label{sec:recovery}

Supporting probes implicate selection and presentation
beyond invocation: 1.7B executes 167/168 supplied-recipe
calls correctly, but fresh-seed selection accuracy changes
from 40.3\% with full tables to 70.0\% with names only.
Distractor, serialization, and padding controls further
show that difficulty is not determined by option count or
input length alone within the tested regimes
(Appendix~\ref{app:controls}).

All nineteen
8B errors in the factorial selection study occur for a
sort target placed first with output name
\texttt{result\_1}, linking errors to a specific interface
configuration. Inert padding within an existing artifact
preserves the H4-trained 1.7B adapter's 100\% accuracy
through roughly 800 tokens and yields 96.25\% at roughly
1,200. This limits a pure length
explanation within that regime, without identifying an
internal mechanism or excluding length effects elsewhere.

Improved local agreement does not ensure completion on
branching tasks. Untuned 8B H16 post-stage-four canonical
accuracy rises from 38.8\% to 77.5\% under compaction,
but success changes from 1/20 to 3/20 ($p=.625$); the H8
completion contrast is also inconclusive ($p=.070$).
Moreover, 65 audited canonical mismatches are legal actions
that increase progress. Canonical agreement and verified
task correctness must therefore remain distinct
(Appendix~\ref{app:branching}).

Completion can coexist with frequent rejections
(Table~\ref{tab:27b}). Untuned 27B
completes 16/20 full-state H16 tasks despite rejected calls
in every episode, averaging 22.8 calls. Under compact
observations, it completes 20/20 with only two episodes
containing rejections and 16.1 mean calls
(Appendix~\ref{app:27b}). The separately evaluated tuned
adapters make no rejected calls, so their conditional
recovery rate is undefined. This is descriptive evidence:
naturally occurring errors differ across policies, and
untuned/tuned tasks and inference paths differ. These
observations do not establish comparative recovery under a
matched error; the controlled 8B test below evaluates one fixed
rejection.

We also compared post-error completion from this matched state
(Table~\ref{tab:matched-recovery-8b}). The full-trained adapter
completed 50/50 tasks after rejection under both views, with no
further rejected calls. The compact-trained adapter completed
49/50 under compact observations but 0/50 under full observations;
all 50 compact-trained full-view episodes incurred another
rejection and used the remaining call budget. Thus, this contrast
cannot be attributed to differences in the initial error or
remaining budget. It remains a result from one adapter pair,
one fixed rejection after four actions, and one task family.
Because compact observations retain evaluator-selected inputs,
the result does not isolate artifact-content masking from
relevance information. The no-injection baseline follows
model-generated states, so it does not estimate the causal effect
of injecting the rejection.

\begin{table}[t]
\centering
\small
\setlength{\tabcolsep}{4pt}
\caption{Qwen3-8B H16 matched-error evaluation (one adapter pair, 50 tasks per cell). The same rejected call follows a verified four-action prefix, leaving 27 calls of the 32-call budget. Further rejections count episodes with at least one additional rejected call; calls are mean post-error model calls.}
\label{tab:matched-recovery-8b}
\begin{tabular}{lrrr}
\toprule
Train $\to$ test & Success & Further rejects & Calls\\
\midrule
Full $\to$ full & 50/50 & 0/50 & 12.00\\
Full $\to$ compact & 50/50 & 0/50 & 12.00\\
Compact $\to$ full & 0/50 & 50/50 & 27.00\\
Compact $\to$ compact & 49/50 & 1/50 & 12.30\\
\bottomrule
\end{tabular}
\end{table}

At H16, no full-view task has canonical agreement at every
correct-prefix probe, yet sixteen autonomous runs complete
within the 32-call budget. Free-run success permits retries
after rejection, whereas clean oracle-path agreement requires
a correct prediction at every independently restored state.

\begin{table}[t]
\centering
\small
\setlength{\tabcolsep}{4pt}
\caption{Untuned Qwen3.8-27B on twenty paired tasks per horizon.
Clean oracle counts tasks with canonical agreement at every
separately restored correct-prefix probe; it is not an
error-free rollout count. Recovery counts eventual successes
among free-run episodes containing a rejected call. Mean calls
include all episodes.}
\label{tab:27b}
\begin{tabular}{llrrrrr}
\toprule
View & $H$ & \shortstack{Post-4\\oracle (\%)} & \shortstack{Clean\\oracle} & \shortstack{Free-run\\success} & Recovery & \shortstack{Mean\\calls}\\
\midrule
Full & 8 & 92.50 & 9/20 & 19/20 & 10/11 & 8.9\\
Compact & 8 & 100.00 & 19/20 & 20/20 & 1/1 & 8.1\\
Full & 16 & 82.08 & 0/20 & 16/20 & 16/20 & 22.8\\
Compact & 16 & 99.17 & 18/20 & 20/20 & 2/2 & 16.1\\
\bottomrule
\end{tabular}
\end{table}

\subsection{External evaluation on ScienceWorld}
\label{sec:scienceworld}

To examine these diagnostics outside the generated table workflows, we
fine-tune Qwen3.8-27B on sampled ScienceWorld trajectories from five task
families \citep{wang2022scienceworld}. The study compares local decisions with
autonomous execution across reference-path lengths and state views. It uses
three training seeds and keeps checkpoint-selection DEV states, held-out
official TRAIN/DEV workflows, and a separate official-test pilot distinct.
Reference paths define short (at most 24 actions), medium (25--40), and long
(41--100) bins; length covaries with task family and is not independently
manipulated. The ScienceWorld observations are constructed differently from
the artifact views in Equation~\ref{eq:compact}
(Appendix~\ref{app:scienceworld}).

Matched-view local agreement
rises from approximately 35\% to 79\% on checkpoint-selection
DEV states. Short-only training yields 26/30 completions
on held-out short workflows but 5/30 on long workflows
(ten tasks per bin, evaluated under three seeds).
Natural length covaries with task semantics, and training
diversity and checkpoint selection differ across conditions;
this is not an isolated horizon effect
(Appendix~\ref{app:scienceworld}).

On ScienceWorld, compact-state training improves local
action agreement under both views, reaching 78.64\% on
compact and 76.89\% on full DEV observations. This evaluation
uses a different model, task structure, and observation
construction from the 1.7B artifact crossover.

In the separate official-test pilot, full-trained checkpoints complete
7/30 rollouts under full observations and 0/30 under
compact observations; compact-trained checkpoints complete
8/30 and 6/30, respectively (Table~\ref{tab:scienceworldcross}).
Each cell evaluates ten tasks under three training seeds.
The local DEV probes use different tasks, so their accuracy
cannot be paired with these completion outcomes.

\begin{table}[t]
\centering
\small
\caption{Paired ScienceWorld evaluation on 30 held-out official TRAIN/DEV workflows. Three training seeds yield 3,126 reference-state predictions and 90 rollouts per condition; completion requires score 100 within 100 calls. The final column excludes the parser rejection defined in Appendix~\ref{app:scienceworld}, not all semantically invalid actions.}
\label{tab:scienceworldpaired}
\begin{tabular}{lrrr}
\toprule
Training & Local (\%) & Completion & \shortstack{No parser\\rejection (\%)}\\
\midrule
Short-only & 69.83 & 46/90 & 95.96\\
Mixed-horizon & 81.61 & 27/90 & 89.58\\
\bottomrule
\end{tabular}
\end{table}

On the same 30 held-out workflows, mixed-horizon training
yields higher reference-action agreement than short-only
training but fewer autonomous completions
(Table~\ref{tab:scienceworldpaired}). Each checkpoint is
probed and rolled out on the same tasks and view; local
accuracy averages decisions, whereas completion averages
episodes. The training condition with higher local agreement
therefore need not have higher completion. In the separate
official-test pilot, 11/63 episodes with parser rejection
eventually succeed (Table~\ref{tab:scienceworldrecovery}).
These are recoveries from naturally occurring errors, not
outcomes under matched injected failures.

\section{Interpretation and Limitations}
\label{sec:discussion}

\paragraph{Interpretation.}
The clearest replicated result is an endpoint interaction between training and
evaluation views. At H16, all three compact-trained 1.7B adapters complete the
50 common workflows in compact view and none in full view; the full-trained
adapters vary across seeds. The decline in full-view correct-prefix accuracy
seen during one compact-trained learning curve is not itself consistent across
seeds. Reference-independent masking and content-dependent paired choices show
that the asymmetry is not restricted to a mask selected from the gold next
action. They do not show that compact training is universally preferable:
full-view training remains near chance on the pivotal paired choices under the
tested budget.

\paragraph{What the controls identify.}
The matched-cue control holds the explicit input-name hint fixed and changes
which artifact contents are displayed. It narrows, but does not eliminate, the
explanation based on evaluator-supplied relevance information. Recency and
state-load controls likewise show that displayed content matters without
separating its amount from position, salience, or prompt length. The
paired-choice task makes both calls executable, but uses independent stage
inputs; it tests content-dependent selection rather than dependency
composition. Gold-action SFT supplies successful actions and immediate
targets, so the experiments do not independently test exploration or temporal
credit assignment.

\paragraph{Generalization and execution.}
The 8B crossover and matched-error probe each use one adapter pair. The 27B
runs use different adaptation, exposure, and inference settings from the
smaller models; their horizon-matched results do not measure short-to-long or
cross-view training transfer. Untuned and tuned 27B evaluations also use
different tasks and inference paths. Successful free runs can include rejected
calls, while a clean correct-prefix path tests a different property; the
controlled 8B probe standardizes one rejection but does not estimate the
causal effect of error feedback. ScienceWorld extends the evaluation to a
public simulator, but its natural horizon covaries with task semantics and its
state views differ from the artifact intervention. The table tasks supply
recipes and reject incorrect calls without corrupting state, leaving memory
reconstruction, unrestricted tool discovery, and destructive-error repair
outside the present setting.

\section{Conclusion}

Supervised agents in our controlled tool workflows can complete tasks at their
training horizon yet fail when the same task state is presented in another
view. The H16 endpoint crossover replicates this compact-to-full transfer
failure across three 1.7B seed pairs, and a separate content-dependent choice
task exhibits it without target-selected masking. These results distinguish
learning the supplied actions from applying them under a changed observation.
The full-view learning-curve decline seen in one seed does not generalize
uniformly, and full-view training does not solve every task family under the
tested budget.

The controls constrain the explanation rather than reduce it to a single
cause: explicit relevance cues help, while masking changes content visibility,
salience, and input length together. Correct-prefix agreement and completion
also diverge because autonomous runs can encounter and recover from errors.
Testing broader workflows with matched observation and feedback interventions
is needed to establish when representation-sensitive transfer becomes a
general limit on long-horizon agents.

\subsection*{AI use statement}
AI tools assisted with research-code development and manuscript language editing, including restructuring draft prose. Reported numerical evidence comes from experiment artifacts and run summaries; AI output is not treated as an experimental observation. The authors are responsible for the analysis, interpretation, and final text.

\subsection*{Ethics statement}
The controlled workflow dataset is procedurally generated and contains no
personal or human-subject data. ScienceWorld uses public simulated tasks. We
disclose evaluator assistance and the restricted error dynamics so that the
results are not mistaken for evidence of autonomous tool discovery or recovery
from destructive actions.

\subsection*{Reproducibility statement}
Task definitions, split manifests, training configurations, checkpoint
identities, and per-action evaluation records are retained for the controlled
studies. The appendix specifies preprocessing, sampling, metrics, and the
provenance and limitations of the separately collected ScienceWorld results.
We will release the generated workflow dataset, including task definitions,
supervised examples, and evaluation split specifications.

\bibliographystyle{iclr2027_conference}
\bibliography{references}

\clearpage
\appendix

\section{Implementation and Protocol Details}
\label{app:implementation}

\subsection{Workflow construction and interaction}
\label{app:workflowdetails}

A task is a directed acyclic graph with $H$ required operation
nodes and edges representing artifact dependencies.
We define the workflow horizon $H$ as the number of required
operations, rather than the dependency depth or the number
of calls made during execution.
The main task family consists of chains with
$H\in\{4,8,16\}$, and an additional family uses shallow
branching. Each operation is specified by an action recipe
containing a tool, input artifact identifiers, an output
identifier, and parameters. The main generator repeats four
unary table transformations: mapping a numeric column,
sorting rows, filtering rows, and selecting columns.
Tables contain bounded values, and deterministic tools
perform the transformations exactly.

A verifier checks artifact contents and valid dependency
ancestry. Success requires verified completion of all
required nodes; producing a final artifact without the
required verified dependencies is insufficient.

At each interaction, the agent receives an observation of
the current artifact state, visible recipes, and the
remaining action budget. It returns a JSON action containing
\texttt{tool}, \texttt{inputs}, \texttt{output}, and
\texttt{params}. The environment rejects malformed actions,
actions that do not match a task recipe, repeated outputs,
and calls with missing prerequisites. A rejected call
consumes one unit of the action budget and leaves the
artifacts unchanged; the next prompt includes the tool
error. We define observed recovery as eventual verified
task completion after at least one rejected call.
Because rejection leaves artifacts unchanged, recovery
does not require repairing corrupted artifacts.
All main autonomous evaluations allow at most $2H$ calls,
including rejected calls.

The main horizon experiments supply two candidate recipes:
the evaluator-selected next recipe and one distractor from
the same task. Distractor selection and target position are
deterministic functions of the task seed and stage.
The ordinary action prompt does not explicitly label the
target, although including it among two candidates already
provides assistance. On a chain, recipe dependencies and
artifact availability identify the unique executable
unfinished operation. The model is not required to perform
the table transformations itself. These tasks therefore
evaluate state-conditioned recipe selection and tool
invocation through a controlled interface.

The evaluator-selected node defines the canonical target
used for action-agreement scoring and determines which
input artifacts retain visible contents in compact observations
(Section~\ref{sec:views}). Execution validity is determined
by the environment's recipe and dependency checks.
On branching tasks, other ready nodes can also be valid
choices, so disagreement with the canonical target does
not itself constitute an invalid action.

\subsection{Observation construction}
\label{app:viewdetails}

At interaction $t$, let $A_t$ map available artifact names
to their contents. The \Full{} view displays the contents
of all available artifacts. It represents the current
accumulated artifact state, rather than a transcript of all preceding model
responses. Prompts are reconstructed at each interaction,
with the previous tool error included when present.

For the \Compact{} view, let $I_t^*$ denote the input
identifiers of the reference next action: the gold action
during SFT and the evaluator-selected next action during
evaluation. For each available artifact name
$k\in\operatorname{dom}(A_t)$, the displayed map is
given by Equation~\ref{eq:compact}.
All available artifact names remain visible; the
\texttt{available} marker indicates that an artifact exists
but its contents are hidden. For paired observations of the
same state, the candidate recipes and action requirements
are unchanged. Both views use the same transition rules
and verification procedure. On branching tasks, contents remain
visible for the inputs of one canonical ready node, even when
other ready nodes also provide valid action choices.

This intervention changes both the amount of displayed
artifact content and the salience of the selected inputs.
Because retention depends on the reference action,
the retention pattern can itself provide relevance
information. The full--compact contrast therefore measures
the effect of this combined observation intervention.
Attributing its effects specifically to reduced distraction
from displayed artifact content would require a control that
matches relevance information.

\subsection{Training}

Gold-prefix examples are generated by executing the required action sequence and recording the observation before each action. In the 1.7B and 8B horizon experiments, each training dataset contributes 800 action examples. The count is fixed at the action level, so distinct training workflows number 200 at H4, 100 at H8, and 50 at H16. Eighty action examples are supplied for development evaluation where enabled. The compact dataset is obtained from the full dataset by replacing the contents of artifacts not required by the reference action with \texttt{available}; target completions and task identifiers are preserved exactly. No RL update or trajectory-level reward optimization is performed. The separate 27B protocol and its learning results appear below and in Appendix~\ref{app:27b-learning}.

\begin{table}[h]
\centering
\caption{Training settings for the 1.7B and 8B horizon and crossover experiments. The original H4 transfer adapter used maximum sequence length 1,024; the separate 1.7B horizon-learning runs use 2,048.}
\label{tab:hyperparameters}
\begin{tabular}{p{0.43\linewidth}p{0.48\linewidth}}
\toprule
Setting & Value\\
\midrule
Base checkpoints & Qwen3-1.7B; Qwen3-8B\\
Supervision & Gold JSON action, completion-only loss\\
Training examples / optimizer steps & 800 actions / 100 steps\\
Microbatch / gradient accumulation & 1 / 8\\
Learning rate / schedule & $2\times10^{-4}$ / linear\\
LoRA rank / alpha / dropout & 16 / 32 / 0.05\\
LoRA target modules & Attention q/k/v/o and MLP gate/up/down projections\\
1.7B optimizer & AdamW\\
8B optimizer / quantization & Paged AdamW 8-bit / NF4, double quantization\\
Compute & BF16; gradient checkpointing\\
Maximum length, horizon runs & 1.7B: 2,048; 8B H4/H8/H16: 640/1,024/1,536\\
Maximum length, H16 crossover & 2,048 (1.7B); 1,536 (8B)\\
8B H16 crossover adapter seed & 20260920\\
Main training and data seed & 20260910\\
H4 replication training seeds & 20260910--20260914\\
\bottomrule
\end{tabular}
\end{table}

Equal action counts do not imply equal input-token exposure or
compute; checkpoints at the same step match optimizer updates,
not wall-clock time. Interrupted 1.7B and 8B training processes
resumed from saved training state. The retained 27B logs show
continuous runs without resume. Process interruptions are excluded
from the agent's in-environment error and recovery measures;
their causes were not verified.

\begin{table*}[t]
\centering
\small
\renewcommand{\arraystretch}{1.15}
\setlength{\tabcolsep}{5pt}

\begin{tabularx}{\linewidth}{
    @{}l>{\raggedright\arraybackslash}Xrr@{}
}
\toprule
Model & Trainer-visible data size
      & Curve tasks & Endpoint tasks \\
\midrule
Qwen3-1.7B
    & 800 action examples
    & 10 & 20 \\
Qwen3-8B
    & 800 action examples
    & 10 & 20 \\
Qwen3.8-27B
    & 800 (H4), 1,600 (H8), and 3,200 (H16) action examples
    & 20 & 20 \\
\bottomrule
\end{tabularx}

\caption{Training and evaluation setup. Data sizes count
examples available to each trainer, not presentations or
distinct examples consumed. Task counts are per condition;
curve counts refer to optimizer steps 20, 40, 60, 80, and 100.
Table~\ref{tab:27b-exposure} separates the 27B data and exposure counts.}
\label{tab:designsetup}
\end{table*}

\paragraph{Source data and training exposure.}
Table~\ref{tab:designsetup} counts examples available to the
trainer. We distinguish these from distinct actions consumed
and repeated presentations contributing to the final checkpoint.
The 1.7B and 8B runs use 800 action examples and
$100\times8=800$ presentations. The 27B runs have different
source sizes and, at H16, a larger optimizer batch
(Table~\ref{tab:27b-exposure}); equal update counts do not
match training exposure.

\begin{table}[h]
\centering
\small
\caption{Audited 27B training exposure. Source and trainer-visible
row counts coincide. ``Distinct'' counts consumed source actions,
not presentations. All source files contain 200 workflows;
H8 covers only some actions from most workflows. Its consumed
indices and workflow coverage are reconstructed from the sampler
and checked against the logged token total.}
\label{tab:27b-exposure}
\begin{tabular}{rrrrrrr}
\toprule
$H$ & Rows & Steps & Examples/update & Presentations & Distinct & Workflows\\
\midrule
4 & 800 & 100 & 8 & 800 & 800 & 200\\
8 & 1,600 & 100 & 8 & 800 & 800 & 200\\
16 & 3,200 & 100 & 64 & 6,400 & 3,200 & 200\\
\bottomrule
\end{tabular}
\end{table}

\paragraph{27B training and preprocessing.}
Three separate Qwen3.8-27B adapters use full-state H4, H8,
or H16 examples from 200 workflows per horizon. Each JSONL
row supplies one gold-prefix observation and its gold action.
All 800/1,600/3,200 rows entered preprocessing: no subset
reduced these counts. Retained training-time bytecode shows
that an optional limit would select a file prefix before
sampling; the exact launch arguments were not preserved.
The saved configuration uses completion-only loss with no
packing, row aggregation, or internal train/dev split.
Retokenization with the saved tokenizer yields maximum
lengths of 616/906/1,505 tokens, below the 1,536-token cap,
and 30--35 supervised completion tokens per row. The
prompt-alignment warnings did not remove rows or eliminate
their supervised tokens. All 3,200 H16 rows were retained.

All runs use NF4 double quantization with BF16 compute,
LoRA rank 16, alpha 32, dropout 0.05, attention q/k/v/o and
MLP gate/up/down targets, paged 8-bit AdamW, learning rate
$2\times10^{-4}$, and 100 optimizer updates. Training and
data seeds are 20260910. A seeded random sampler draws
without replacement within an epoch and uses a new
permutation for the next epoch. H4 consumes all 800 actions
once; H8 consumes 800 of 1,600 actions, representing all
200 workflows; H16 consumes all 3,200 actions twice.
Reconstructed token totals match the logged 409,788,
493,030, and 5,207,038 tokens, respectively. In particular,
H8's 200 represented workflows are not 200 complete
supervised trajectories.

Per-device batch size is one and gradient accumulation is
eight in each run. H4/H8 save a GPU-count batch factor of
one; H16 saves eight in a single non-DDP process, giving
64 examples per update, corroborated by epoch and token
accounting. The quantized H16 model was loaded on GPU~0;
active compute replicas cannot be recovered from the
retained artifacts. We therefore report the verified batch
and exposure, without asserting eight-way data parallelism.
The retained logs show one continuous run to step 100 per
horizon and no resume; presentations from any unretained
failed attempts are unknown and excluded. These records
verify training execution and exposure. The evaluation
trajectories require separate verification.

\paragraph{27B adapter loading.}
The training and local evaluation wrappers place the Qwen3.8 decoder under different module paths. Before checkpoint evaluation, 512 LoRA tensor keys were renamed for the text-only loader without changing their values; the corrected adapters loaded with no missing or unexpected keys, and a smoke test with a corrected checkpoint changed behavior relative to the base model. The correction does not affect the untuned vLLM run. The complete evaluation bundle, including model and adapter identities and raw trajectories, has not yet been consolidated.

\subsection{Comparison design}
\label{app:comparisondesign}

We construct horizon-matched learning curves by evaluating
successive checkpoints from each training run on the same
held-out tasks at the training horizon and under a fixed
state view. Horizon-transfer evaluations hold the checkpoint
and state view fixed while varying the evaluation horizon,
including workflows longer than those used for training.
Comparisons between separately trained adapters evaluated
at their respective training horizons characterize
performance under matched supervision; they do not by
themselves establish transfer to longer workflows.

The H16 training--evaluation representation crossover uses
the 1.7B model. Full and compact training datasets contain
the same task instances, gold prefixes, and target actions;
only the displayed artifact view changes. Each trained
adapter is evaluated under both views on common held-out
tasks. The full-trained/full-evaluated arm reuses the
corresponding H16 horizon-matched experiment and does not
constitute an independent replication. The initial H16 comparison used one
seed; two additional paired full/compact training seeds were subsequently
evaluated on 50 common held-out workflows. The separate five-seed H4
replication does not estimate H16 training-run variation.

The artifact-workflow 27B checkpoint curves use full-state oracle prefixes,
and final evaluations cross full and compact views.
Because model generation, adaptation method, and effective
data exposure differ, comparisons across model sizes describe
the evaluated configurations and do not isolate the effect
of parameter count.

Supplied-recipe probes measure tool-call generation when the
required action is given. Comparing these with recipe-selection
probes tests whether successful invocation is sufficient for
correct selection in the evaluated states.

\subsection{Evaluation and data partitions}

Local 1.7B and 8B evaluation uses greedy decoding with thinking disabled, a limit of 160 generated tokens per call, and BF16 model loading. Thus, 8B uses quantized training and BF16 evaluation. The primary local device is an RTX 3090. Untuned Qwen3.8-27B uses BF16 vLLM serving on an A100 cluster, whereas the fine-tuned 27B adapters are evaluated through a local NF4 base with corrected LoRA loading; the H16 evaluation reports a single L20Y GPU. These inference paths are not matched before and after tuning. The action budget is $2H$ for autonomous evaluation; an oracle-prefix sweep probes $H$ correct states. A correct action must match the required recipe and be legal in the environment.

For the 1.7B and 8B SFT studies, training tasks start at seed 60000, development tasks at 61000, and evaluation tasks use 62000--62019, with generator split labels recorded in manifests. Their learning curves use 62000--62009. The fresh H16 compact confirmation uses 63000--63019. The matched untuned-model protocol, including untuned 27B, uses 64000--64019. Each 27B horizon-matched checkpoint curve and final evaluation uses test-split seeds 62000--62019, disjoint from its reported training seeds 60000--60199 and development seeds 61000--61019. The 27B curves use full-state oracle prefixes; step-100 final evaluations cross full/compact observations and oracle/free-run modes. Exact task fingerprints and generator settings still require consolidation with the cluster manifests; numerical seeds alone do not fully identify a task. Full/compact comparisons within a run pair the same tasks, whereas untuned/tuned comparisons do not.

\subsection{Metrics and uncertainty}
\label{app:metricdetails}
\emph{Oracle-prefix accuracy} is the fraction of separately
probed predictions that agree with the canonical next
action. Before each prediction, the environment is restored
to the corresponding gold-prefix state.
\emph{Clean oracle-path rate} is the fraction of tasks for
which all such predictions agree with their canonical
actions. Both metrics evaluate decisions on correct
intermediate states; clean oracle-path rate is not an
autonomous success rate.

\emph{Free-run success} is the fraction of episodes that
satisfy the exact task verifier at termination, with the
model's actions determining subsequent progress.
Let $C_T$ denote the set of required nodes whose outputs
and dependency ancestry have been verified when an episode
ends. We define \emph{verified final progress} as $|C_T|/H$,
counting verified nodes regardless of their execution order.
Mean calls includes all attempted calls, including rejected
calls, and is averaged across all episodes, successful
and unsuccessful.

For $N$ evaluation episodes, let $S_i$ indicate verified
task success and let $E_i$ indicate whether at least one
call was rejected. Episode-level rejected-call incidence
is $N^{-1}\sum_{i=1}^{N} E_i$. We define
\emph{observed recovery} as
\begin{equation}
\widehat{R} =
\frac{\sum_{i=1}^{N} S_i E_i}
{\sum_{i=1}^{N} E_i},
\qquad \sum_{i=1}^{N} E_i > 0.
\label{eq:observedrecovery}
\end{equation}
This is eventual verified success conditional on an episode
containing a rejected call. It is undefined when no episode
contains a rejection. Error-free completion requires both
$S_i=1$ and $E_i=0$. On branching tasks, canonical action
disagreement and call rejection are scored separately,
since a noncanonical action can still be valid.

\paragraph{Sample sizes, pairing, and uncertainty.}
The original endpoint evaluations use 20 tasks per condition
for each model. The follow-up training pilot in
Section~\ref{sec:augmentation} uses ten development tasks
per condition and is reported separately. The ScienceWorld
studies in Section~\ref{sec:scienceworld} use separate
sampled task sets with three training seeds; their split
roles and denominators are specified there. Full--compact
contrasts within an evaluation run
use the same task seeds. The shared horizon-learning
comparison evaluates optimizer steps 20, 40, 60, 80, and 100
(Table~\ref{tab:horizoncurves}). At these checkpoints, the
1.7B and 8B curves use a ten-task subset of their endpoint
seeds, whereas the 27B curves use twenty tasks. Additional
27B evaluations at optimizer steps 5 and 10 characterize
early learning.

For comparisons of fixed checkpoints, statistical analyses
use tasks as the sampling unit. Where raw paired outcomes
are available, comparisons preserve task pairing rather
than treating actions within a task as independent
observations. The original two-adapter H16 crossover intervals use 20,000
paired task-bootstrap resamples. The corresponding two-sided
sign-flip tests are exploratory and unadjusted for multiple
comparisons. These intervals and tests condition on the
trained adapters and characterize variation across tasks;
they do not assess variation across training seeds. The later
three-pair replication is summarized separately with per-seed
interactions.

Untuned 27B full--compact comparisons are paired within
their own evaluation tasks. Comparisons with the separately
evaluated 27B adapters involve different task seeds and
inference paths. We therefore report these comparisons
descriptively and do not interpret the differences as
controlled estimates of the effect of SFT. Development
experiments informed later protocols; only experiments
explicitly designated as fresh-seed confirmations should
be interpreted as such. Appendix~\ref{app:statistics}
provides further statistical details.

\subsection{Observation masking and execution}

Section~\ref{sec:views} illustrates how masking changes the
displayed contents while preserving the underlying artifacts.
A recipe whose output already exists is rejected. After an
invalid choice, the environment returns an error and reduces
the remaining budget; it does not execute the gold action
in a free run.

\subsection{\texorpdfstring{Horizon and accumulated state}{Horizon and accumulated state}}
\label{app:horizon-state}

The fixed-checkpoint control uses the H4 full-view
Qwen3-1.7B adapter at optimizer step 100 (training seed
20260910) and 20 test seeds, 127000--127019. At H8, added
sources are real but have no recipe dependencies. For a seed,
all conditions share the original operation nodes, target
rank, candidate pair, and $2H$ budget. Tables with 4$\times$16
versus 8$\times$8 rows and 4$\times$32 versus 8$\times$16
rows match total distractor rows, not exact token counts.
The horizon control uses H4/H8/H16 prefixes of each seed's
H16 chain, with shared early operations and candidate-pair
construction. Its Recent-2 view selects contents by successful
artifact-creation history without the reference action; all
artifact names and the two candidate recipes remain visible.

\begin{table}[t]
\centering
\small
\caption{Fixed-H8 state-load control on 20 paired test
tasks. Each condition has 160 separately restored correct-prefix
decisions and 20 free runs. Tokens are mean correct-prefix
input lengths. The largest load does not reproduce the
original horizon decline; effects across loads are nonmonotone.}
\label{tab:horizon-state-load}
\begin{tabular}{lrrr}
\toprule
Extra artifacts $\times$ rows & Tokens & Oracle actions & Free-run success\\
\midrule
0 & 542 & 138/160 & 8/20\\
4 $\times$ 8 & 802 & 130/160 & 4/20\\
4 $\times$ 16 & 1,018 & 135/160 & 6/20\\
4 $\times$ 32 & 1,466 & 137/160 & 8/20\\
8 $\times$ 8 & 1,062 & 131/160 & 4/20\\
8 $\times$ 16 & 1,494 & 135/160 & 7/20\\
8 $\times$ 32 & 2,390 & 136/160 & 7/20\\
\bottomrule
\end{tabular}
\end{table}

\begin{table}[t]
\centering
\small
\caption{Prefix-matched horizon control on 20 tasks per
cell using the same H4-trained adapter. Recent-2 limits shown
contents to at most two artifacts but retains all names.
Correct-prefix accuracy and free-run completion measure
different outcomes; the latter uses $2H$ calls.}
\label{tab:horizon-state-horizon}
\begin{tabular}{rlrrr}
\toprule
$H$ & View & Tokens & Oracle actions & Free-run success\\
\midrule
4 & Full & 440 & 79/80 & 19/20\\
4 & Recent-2 & 396 & 79/80 & 19/20\\
8 & Full & 546 & 139/160 & 8/20\\
8 & Recent-2 & 405 & 151/160 & 13/20\\
16 & Full & 752 & 230/320 & 0/20\\
16 & Recent-2 & 430 & 294/320 & 4/20\\
\bottomrule
\end{tabular}
\end{table}

In successive four-action blocks on H16 prefixes,
full-view correct-prefix agreement is 79/80, 60/80, 49/80,
and 42/80; Recent-2 gives 78/80, 73/80, 73/80, and 70/80.
At H16 the Recent-2 view yields more correct actions for
19/20 paired task seeds (one tie), but only 4/20 free runs
complete. Under the same Recent-2 view, completion decreases
from 19/20 at H4 to 4/20 at H16 as the number of required
calls grows, while mean input length changes from 396 to
430 tokens. This is inconsistent with a sufficient explanation
based only on accumulated displayed table contents. It does
not isolate a pure horizon effect: names and output identifiers
still accumulate, the call budget scales with $H$, and recency
may provide relevance information in a chain. Results use one
adapter and 20 task seeds.

\subsection{\texorpdfstring{Paired content-dependent choices}{Paired content-dependent choices}}
\label{app:content-choice}

The H16 paired-choice data comprise 100 training pairs
(200 workflows and 3,200 gold-action examples per view), 20
development pairs, and 50 held-out test pairs (seeds
72000--72049). One stage in each pair flips one table cell
between $-1$ and $+1$; recipe order depends on seed and stage,
not on the target. The active input is explicit and shared by
both candidates. Full observations show all available artifact
contents, while compact observations show only the active
input contents and retain every artifact name. The two exact
candidate JSON completions have equal tokenizer length on
all pivotal test probes. The maximum measured full-view
training sequence length is 1,396 tokens, below the 2,048-token
limit; no training example is sequence-truncated.
Each of six LoRA runs uses 100 updates, effective batch size
eight, and one of three paired training seeds (20261001--20261003).
The greedy rollout subset is fixed before evaluation as the
first ten test pairs, with a 32-call cap. A goal-wrong but valid
call executes, advances the stage, and makes final success
impossible; invalid calls consume budget without advancing.

\begin{table}[h]
\centering
\small
\caption{Per-seed H16 paired-choice results. Pivotal
correct counts are out of 100 choices, both-variant counts
out of 50 pairs, and free-run success out of 20 tasks. All
views use the same frozen test pairs.}
\label{tab:content-choice-seeds}
\begin{tabular}{llrrr}
\toprule
Seed & Train $\to$ test & Pivotal & Both & Free run\\
\midrule
20261001 & Full $\to$ full & 52 & 2 & 0\\
20261001 & Full $\to$ compact & 59 & 9 & 0\\
20261001 & Compact $\to$ full & 59 & 9 & 0\\
20261001 & Compact $\to$ compact & 100 & 50 & 20\\
20261002 & Full $\to$ full & 49 & 3 & 0\\
20261002 & Full $\to$ compact & 61 & 12 & 0\\
20261002 & Compact $\to$ full & 59 & 9 & 0\\
20261002 & Compact $\to$ compact & 100 & 50 & 20\\
20261003 & Full $\to$ full & 50 & 0 & 0\\
20261003 & Full $\to$ compact & 53 & 3 & 0\\
20261003 & Compact $\to$ full & 59 & 9 & 0\\
20261003 & Compact $\to$ compact & 100 & 50 & 20\\
\bottomrule
\end{tabular}
\end{table}

\section{Statistical Analyses and Crossover Results}
\label{app:statistics}

\begin{table}[t]
\centering
\caption{Original-seed H16 training crossover for Qwen3-1.7B on its initial 20-task endpoint set. The two adapters use the same task/action examples and training seed. Oracle accuracy uses 320 correct-prefix decisions; free-run success uses paired tasks. The later three-seed evaluation is reported in Table~\ref{tab:crossover-seeds}.}
\label{tab:crossover}
\begin{tabular}{llrrr}
\toprule
Train view & Test view & Oracle (\%) & Free-run success & Mean calls\\
\midrule
Full & Full & 99.1 & 17/20 & 18.40\\
Full & Compact & 96.9 & 15/20 & 20.40\\
Compact & Full & 32.2 & 0/20 & 32.00\\
Compact & Compact & 100.0 & 20/20 & 16.00\\
\bottomrule
\end{tabular}
\end{table}

For the original seed and its 20-task endpoint set, let $S_{ab}$ denote free-run success when training in view $a$ and evaluating in view $b$. The observed difference of differences is
\begin{equation}
 \Delta=(S_{CC}-S_{CF})-(S_{FC}-S_{FF})
        =(1.00-0.00)-(0.75-0.85)=1.10.
 \label{eq:interaction}
\end{equation}
The paired task-bootstrap 95\% interval is $[0.85,1.35]$, with exploratory two-sided sign-flip $p=1.53\times10^{-5}$. A difference of differences can exceed one even though each success rate lies in $[0,1]$. In contrast, the full-trained adapter's 85\% versus 75\% success difference is inconclusive (paired $p=0.727$). The supported pattern is the large asymmetry in cross-view transfer; the data do not establish that full-state training is generally more robust or that compact evaluation harms full-trained models on average.

For each task seed, the crossover supplies four outcomes, one per train/test cell. We resample whole task seeds, retaining their four outcomes together. The interaction statistic is calculated per seed and then averaged. The paired bootstrap uses 20,000 draws with analysis seed 20260915. Two-sided sign-flip tests evaluate paired contrasts under a sign-exchangeability null; these post hoc crossover comparisons are exploratory and receive no multiplicity correction. Their interpretation is conditional on the two realized adapters.

\paragraph{H16 endpoint replication.}
The original H16 pair (seed 20260910) and two new pairs (seeds
20260926 and 20260927) were evaluated on the same frozen test
workflows, task seeds 126000--126049. Each pair used the first
800 H16 source action examples, representing 50 training
workflows, with 100 optimizer steps and effective batch size
8. Each new run completed uninterrupted; exposure logs verify
800 distinct examples and 800 total presentations. Full and
compact adapters within a seed pair use the same training seed.
The seed-specific completion interactions in
Table~\ref{tab:crossover-seeds} are descriptive. Their mean and
sample standard deviation summarize three seed pairs; we do not
use the 150 seed-by-task records as independent observations.

\begin{table}[t]
\centering
\small
\setlength{\tabcolsep}{3pt}
\caption{Matched-cue and recent-artifact controls at H16 on the same 20 held-out tasks per checkpoint. Each cell reports correct-prefix action agreement (\%) and free-run completions. The explicit cue gives the same evaluator-selected input-artifact names in both cued views.}
\label{tab:relevance-controls}
\begin{tabular}{lccccc}
\toprule
Checkpoint & Full & Full+cued & Compact & Compact+cued & Recent-2\\
\midrule
H4-trained & 66.56; 0/20 & 88.12; 7/20 & 98.12; 17/20 & 99.38; 19/20 & 78.44; 0/20\\
H16 full-trained & 98.75; 19/20 & 99.69; 20/20 & 94.06; 8/20 & 99.38; 19/20 & 96.88; 15/20\\
H16 compact-trained & 37.50; 0/20 & 64.06; 1/20 & 100.00; 20/20 & 100.00; 20/20 & 60.94; 0/20\\
\bottomrule
\end{tabular}
\end{table}

For the H4-trained adapter, 12 paired tasks succeeded only under
compact+cued and none succeeded only under full+cued (exploratory
exact two-sided paired $p=0.00049$, unadjusted across controls).
Adding the explicit cue raised full-view completion from 0/20 to
7/20, while compact+cued completion was 19/20. The target-independent
Recent-2 view improved correct-prefix agreement over full view for
this adapter (78.44\% versus 66.56\%) but did not improve completion
(0/20 in both views). For the original compact-trained H16 adapter,
full+cued completion remained 1/20 while compact+cued was 20/20.
The control set is small and fixed-checkpoint; the relevance cue was
not used during SFT. All conditions retain the evaluator-selected
action among the two candidate recipes, and Recent-2 is specifically
suited to chain recency. These results test selected controls; they
do not establish an unassisted policy or a general masking rule.

\begin{table}[h]
\centering
\caption{Task-conditional crossover contrasts in free-run success. Effects and 95\% paired-bootstrap intervals are in percentage points. The bootstrap interval at the observed 100-point contrast is degenerate because every empirical paired difference is identical; it does not imply zero population uncertainty.}
\label{tab:contrasts}
\begin{tabular}{p{0.42\linewidth}rrr}
\toprule
Contrast & Effect & 95\% interval & $p$\\
\midrule
Compact minus full training, full test & $-85$ & $[-100,-70]$ & $1.53\times10^{-5}$\\
Compact minus full training, compact test & $25$ & $[10,45]$ & $.0625$\\
Compact minus full test, full training & $-10$ & $[-35,15]$ & $.727$\\
Compact minus full test, compact training & $100$ & $[100,100]$ & $1.91\times10^{-6}$\\
Difference of differences & $110$ & $[85,135]$ & $1.53\times10^{-5}$\\
\bottomrule
\end{tabular}
\end{table}

Bootstrap intervals and discrete sign-flip tests use different constructions and can disagree near a decision threshold; for example, the 25-point contrast has an interval excluding zero but a two-sided test of $.0625$. We do not treat that contrast as decisive evidence. Marginal 95\% Wilson intervals are $[0,16.1]\%$ for 0/20, $[53.1,88.8]\%$ for 15/20, $[64.0,94.8]\%$ for 17/20, and $[83.9,100]\%$ for 20/20. Boundary sample outcomes must not be interpreted as certain population success or failure.

Figure~\ref{fig:crossover} shows the original seed pair's learning curves on a ten-task subset. For its compact-trained adapter, compact-view oracle accuracy rises from 90.0\% at step 20 to 100\% at step 40 and remains there, while full-view accuracy falls from 61.9\% to 33.8\% by step 100. Because each probe restores a correct prefix, earlier rollout errors cannot explain this decline. Its direction and size did not replicate uniformly: the step-20-to-100 full-view changes were $-29.4$, $-3.1$, and $+11.9$ percentage points across the three seed pairs. The curve therefore describes the original adapter, not a general learning trajectory. In that pair, the full-trained adapter reaches 100\% compact-view accuracy at steps 40 and 60 and 96.9\% at step 100. These ten-task curve estimates should not be conflated with the fifty-task endpoint evaluations.

\begin{table}[h]
\centering
\caption{Original-seed H16 crossover oracle-prefix accuracy (\%) on ten tasks per checkpoint (160 decisions). F/C denotes full/compact. The same task subset is evaluated at every checkpoint; endpoint results use a larger, separate sample.}
\label{tab:curves}
\begin{tabular}{lrrrrr}
\toprule
Training $\to$ evaluation & Step 20 & 40 & 60 & 80 & 100\\
\midrule
F $\to$ F & 71.875 & 85.625 & 91.250 & 98.750 & 99.375\\
F $\to$ C & 83.125 & 100.000 & 100.000 & 96.875 & 96.875\\
C $\to$ F & 61.875 & 40.000 & 36.875 & 34.375 & 33.750\\
C $\to$ C & 90.000 & 100.000 & 100.000 & 100.000 & 100.000\\
\bottomrule
\end{tabular}
\end{table}

\begin{table}[h]
\centering
\caption{Additional H16 endpoint metrics. Recovery counts include only episodes with an invalid action; a dash denotes no such episodes. Oracle clean-path agreement uses the gold-prefix sweep, not free-run trajectories.}
\label{tab:crossoverextra}
\begin{tabular}{llrrrr}
\toprule
Train & Test & Post-4 oracle (\%) & Oracle clean & Progress (\%) & Recovery\\
\midrule
F & F & 98.750 & 17/20 & 93.750 & 0/3\\
F & C & 95.833 & 12/20 & 94.063 & 3/8\\
C & F & 18.750 & 0/20 & 17.188 & 0/20\\
C & C & 100.000 & 20/20 & 100.000 & ---\\
\bottomrule
\end{tabular}
\end{table}

\subsection{Untuned 27B observation contrasts}

The full and compact views are paired on twenty untuned 27B tasks per horizon. At H16, clean oracle paths change from 0/20 to 18/20 (exploratory exact two-sided sign-flip $p=7.63\times10^{-6}$), while free-run completion changes from 16/20 to 20/20 ($p=.125$). At H8, the corresponding changes are 9/20 to 19/20 ($p=.00635$) and 19/20 to 20/20 ($p=1.00$). These comparisons are exploratory and unadjusted across endpoints. The local-action evidence is stronger than the completion contrast at these sample sizes; a non-significant completion contrast is not evidence of equivalence. The 20/20 tuned endpoints likewise do not establish population-level certainty. Untuned and fine-tuned 27B runs use different task seeds and inference paths, so their apparent before/after changes have no paired test here.

\section{Supporting Controls and Learning Curves}
\label{app:controls}

\subsection{Interface controls}

The fresh-seed local-interface study uses twenty seeds, 10000--10019. Full-table and names-only conditions preserve the action-selection requirement while varying displayed artifact content. Identifier conditions also simplify output generation, so differences across these groups cannot be attributed solely to input representation. The registered local contrasts used one-sided seed-level sign-flip tests with Holm correction across three planned comparisons; this analysis differs from the exploratory two-sided crossover analysis.

\begin{table}[h]
\centering
\caption{Fresh-seed local-interface study with Qwen3-1.7B. The first four rows contain 16 probes per seed. Input-token counts are reported study means, not context-length limits.}
\label{tab:interfaces}
\begin{tabular}{lrrr}
\toprule
Interface & Correct / probes & Accuracy (\%) & Mean input tokens\\
\midrule
Full tables, full action output & 129/320 & 40.31 & 607\\
Names only, full action output & 224/320 & 70.00 & 374\\
Recipe IDs with full fields & 253/320 & 79.06 & 205\\
Output-only recipe IDs & 314/320 & 98.13 & 124\\
One-distractor configuration & 345/480 & 71.88 & 554.5\\
\bottomrule
\end{tabular}
\end{table}

The names-only minus full-table difference is 29.69 percentage points (paired task-bootstrap interval $[23.13,35.31]$, Holm-adjusted $p=7.63\times10^{-6}$). The output-only ID minus full-field ID difference is 19.06 points ($[16.25,21.56]$, adjusted $p=2.86\times10^{-6}$). These establish sizable interface effects in the tested probe distribution. The one-distractor row has a different probe layout; a later strictly matched distractor experiment is therefore used for the option-count claim in the main text.

In that matched study, ten seeds 30000--30009 supply 240 probes per model and distractor condition. Increasing distractors from one to three changes 1.7B accuracy from 173/240 to 107/240 and 8B accuracy from 230/240 to 240/240. The one-minus-three-distractor difference is $+27.5$ points for 1.7B, with a paired-bootstrap interval $[23.33,32.50]$ and one-sided $p=.00098$. The observed 8B difference has the opposite sign ($-4.17$ points). A subsequent factorial study uses ten seeds 40000--40009, twelve ordered pairs of distinct tools, two target ranks, and two output-label assignments, giving 480 probes per model. Aggregate accuracy is 73.54\% for 1.7B and 96.04\% for 8B. Of the latter model's 19 errors, ten involve a map distractor, eight a filter distractor, and one a select distractor; all share the target-sort, first-position, \texttt{result\_1} configuration.

Untuned 8B reaches 160/160 in both full-table and names-only conditions of an earlier local replication. This local ceiling does not conflict with poor full-state trajectory performance: the later horizon protocol changes the state distribution, and a task requires many decisions rather than a single successful probe.

\subsection{Padding and development studies}

The controlled inert-padding study evaluates twenty H4 tasks, four oracle decisions each, for 80 decisions per padding tier. Accuracy is 100\% without padding and at approximately 600 and 800 tokens, 98.75\% at 1,000, and 96.25\% at 1,200. The largest tier's paired comparison is inconclusive ($p=.25$). An earlier padding variant added a new artifact name, producing 75--80\% accuracy; because it changed artifact identity and binding cues, it is a development diagnostic rather than a clean length control.

Other development studies include primitive invocation, target-recipe copying, all/unfinished/executable recipe views, source-versus-intermediate inputs, neutral output naming, and branch-count sweeps. They informed the main controlled protocol but use heterogeneous task counts and evolving prompts. For example, supplied target recipes yield 167/168 correct matched primitive actions; a small workflow pilot improves from 0/6 under all or unfinished recipes to 5/6 when only executable recipes are displayed. Such results motivate selection controls but are not independent confirmatory replications. We retain their provenance rather than combining their numerators into a single accuracy estimate.

\subsection{Chain observation contrasts}
\label{app:chaincontrasts}

Applying Equation~\ref{eq:compact} to the H4-trained 1.7B adapter raises H8 success from 5/20 to 20/20 and H16 success from 0/20 to 18/20 (Table~\ref{tab:rescue}). Post-stage-four oracle accuracy rises from 68.8\% to 100\% at H8 and from 50.4\% to 98.3\% at H16. On a separate set of 20 fresh H16 tasks, full and compact success remain 0/20 and 18/20, respectively. This confirmation uses a fixed adapter: it replicates across tasks, not across training runs.

Untuned 8B exhibits the same direction on chains. Full-state success is 1/20 at H8 and 0/20 at H16; compact-state success is 15/20 and 16/20. Mean calls fall from 15.75 to 10.30 at H8 and from 32.00 to 20.35 at H16. These improvements show that poor full-state execution need not imply that the model cannot perform the required sequence under another observation interface. They do not establish that hiding artifact contents alone causes the improvement, because the compact view also identifies a subset of relevant inputs.

Untuned 27B also shows a paired full--compact gain in correct-prefix action selection: post-stage-four oracle accuracy rises from 92.5\% to 100\% at H8 and from 82.1\% to 99.2\% at H16 (Table~\ref{tab:rescue}). Free-run success changes from 19/20 to 20/20 and from 16/20 to 20/20, respectively; these success differences are inconclusive at twenty tasks (two-sided paired $p=1.00$ and $p=.125$). At H16, clean oracle paths rise from 0/20 to 18/20, while full-state free runs often complete after invalid actions (Appendix~\ref{app:27b}). This difference between oracle paths and eventual success motivates tracking execution errors separately. The 27B checkpoint is from a different model generation and uses a different inference path, so the rows do not estimate an isolated model-size effect. The horizon-matched 27B adapters' ceiling-level full/compact endpoints appear in Table~\ref{tab:27b-learning}.

\subsection{Horizon-matched learning curves}

Untuned Qwen3-1.7B has oracle-prefix accuracies of 42.5\%, 36.9\%, and 35.9\% at H4, H8, and H16, respectively. Its autonomous success is 10\%, 0\%, and 0\%. H4 SFT reaches perfect H4 action accuracy and completion, with this short-task ceiling replicated across five training seeds. Nevertheless, the H4-trained adapter succeeds on only 25\% of H8 tasks and none of the H16 tasks under full-state evaluation. Its oracle accuracy after stage four falls to 68.8\% at H8 and 50.4\% at H16. Because these probes restore the gold prefix, the decline cannot be explained solely by errors made earlier in the same rollout.

Training directly on the evaluation horizon changes this result. Table~\ref{tab:learnability} shows high endpoint success at all three horizons. On the common full-state oracle-prefix checkpoint grid (Table~\ref{tab:horizoncurves}), H16 accuracy for 1.7B rises from 71.9\% at step 20 to 99.4\% at step 100; 8B reaches 99.4\% at step 20, and 27B is at 100\% at every evaluated step from 20 through 100. These observations establish learnability under the supplied supervision, but do not isolate model size or support a controlled sample-efficiency ranking: model generation, LoRA versus QLoRA, training-data exposure, and curve denominators differ. For the 1.7B and 8B runs, fixing 800 action examples gives 200, 100, and 50 training workflows at H4, H8, and H16, respectively. Nor does the horizon intervention separate depth from the number and size of accumulated artifacts.

\begin{table}[h]
\centering
\caption{Full-state oracle-prefix action accuracy (\%) after horizon-matched SFT at the common checkpoints. The 1.7B and 8B rows use ten tasks per point and training seed 20260910; the 27B rows use twenty tasks per point from separate training runs. Each model--horizon row evaluates a distinct adapter. The 27B H16 run presents 64 examples per update; all other rows present eight. Thus equal steps are not exposure matched.}
\label{tab:horizoncurves}
\begin{tabular}{llrrrrr}
\toprule
Model & Horizon & Step 20 & 40 & 60 & 80 & 100\\
\midrule
1.7B & 4 & 90.000 & 100.000 & 100.000 & 100.000 & 100.000\\
1.7B & 8 & 90.000 & 93.750 & 98.750 & 100.000 & 100.000\\
1.7B & 16 & 71.875 & 85.625 & 91.250 & 98.750 & 99.375\\
8B & 4 & 100.000 & 100.000 & 100.000 & 100.000 & 100.000\\
8B & 8 & 100.000 & 100.000 & 100.000 & 100.000 & 100.000\\
8B & 16 & 99.375 & 99.375 & 100.000 & 100.000 & 100.000\\
27B & 4 & 100.000 & 100.000 & 100.000 & 100.000 & 100.000\\
27B & 8 & 100.000 & 100.000 & 100.000 & 100.000 & 100.000\\
27B & 16 & 100.000 & 100.000 & 100.000 & 100.000 & 100.000\\
\bottomrule
\end{tabular}
\end{table}

\subsection{27B step-100 full/compact evaluation}
\label{app:27b-learning}

Separate Qwen3.8-27B adapters were trained on full-state H4, H8, and H16 examples. Their action accuracies at steps 20, 40, 60, 80, and 100 appear alongside the 1.7B and 8B curves in Table~\ref{tab:horizoncurves}. Each 27B curve uses twenty held-out tasks, or $20H$ oracle decisions per checkpoint, and has 20/20 clean paths at every reported checkpoint. Table~\ref{tab:27b-learning} reports the final full/compact evaluation. The common grid is at ceiling from step 20. The earlier H16 probes reach 319/320 correct oracle actions at step 5 and 320/320 at step 10, after 320 and 640 action-example presentations, respectively. The audited 64-example update batch (Table~\ref{tab:27b-exposure}) rules out interpreting these equal-step comparisons as matched-exposure learning curves.

\begin{table}[h]
\centering
\caption{Step-100 evaluation of horizon-matched Qwen3.8-27B adapters on twenty paired test tasks per horizon. Each cell lists Full, Compact in that order; oracle cells count clean gold-prefix paths, and free-run cells count verified completions. Each horizon uses one separately trained full-state adapter.}
\label{tab:27b-learning}
\begin{tabular}{lcc}
\toprule
$H$ & \shortstack{Clean oracle paths\\F, C} & \shortstack{Free-run success\\F, C}\\
\midrule
4 & 20/20, 20/20 & 20/20, 20/20\\
8 & 20/20, 20/20 & 20/20, 20/20\\
16 & 20/20, 20/20 & 20/20, 20/20\\
\bottomrule
\end{tabular}
\end{table}

At step 100, every evaluated free run completes in exactly $H$ calls without an invalid action, and verified final progress is 100\% in both views. Thus conditional recovery has no error-containing denominator, rather than a 100\% recovery rate. The final full/compact crossing yields eighty evaluation episodes per horizon across two views and two modes, but only twenty task seeds per horizon. All adapters were trained in the full view: this table neither measures short-to-long transfer nor a compact-training/full-testing crossover. The all-success endpoint is an observed ceiling on these tasks, not evidence of equivalent behavior across views on unseen distributions.

\section{Execution Errors and Recovery}
\label{app:branching}

\subsection{Branching task metrics}

We evaluate untuned 8B on a shallow branching graph using the same tool environment. The compact view improves post-stage-four canonical oracle accuracy from 36.3\% to 83.8\% at H8 and from 38.8\% to 77.5\% at H16. Autonomous success changes from 1/20 to 7/20 at H8 and from 1/20 to 3/20 at H16 (Table~\ref{tab:branching}). The paired two-sided success tests are inconclusive ($p=0.070$ and $p=0.625$). At H16, mean verified final progress actually decreases from 39.7\% to 34.1\%, also inconclusively. Thus the branching results support local improvement under the assisted view, but not a reliable task-completion rescue across graph structures.

\begin{table}[t]
\centering
\caption{Untuned 8B on shallow branching tasks, 20 paired tasks per cell. Canonical oracle agreement is not order-invariant task correctness. Parentheses give exploratory two-sided paired success-test $p$-values.}
\label{tab:branching}
\begin{tabular}{lccc}
\toprule
Horizon & Post-4 oracle (\%) & Free-run success & Final progress (\%)\\
 & Full $\to$ compact & Full $\to$ compact & Full $\to$ compact\\
\midrule
8 & 36.3 $\to$ 83.8 & 1/20 $\to$ 7/20 ($.070$) & 36.9 $\to$ 51.9\\
16 & 38.8 $\to$ 77.5 & 1/20 $\to$ 3/20 ($.625$) & 39.7 $\to$ 34.1\\
\bottomrule
\end{tabular}
\end{table}

Canonical action agreement is order-sensitive; task verification is not. In shallow graphs, the compacting rule follows a canonical ready node. If the model executes a different ready recipe, it can make useful progress despite disagreeing with the probe target. The raw-action audit finds 13 such actions for full H8, 24 for full H16, 13 for compact H8, and 15 for compact H16. All 65 are legal and increase verified progress. We therefore preserve canonical agreement as a diagnostic label and use invalid-action status to count execution errors.

\begin{table}[h]
\centering
\caption{Corrected branching recovery counts for untuned 8B. Legal alternative-order actions are excluded from invalid-action counts.}
\begin{tabular}{llrrr}
\toprule
View & $H$ & \shortstack{Episodes with\\invalid actions} & \shortstack{Succeeded\\among those} & \shortstack{Legal noncanonical\\actions}\\
\midrule
Full & 8 & 20/20 & 1/20 & 13\\
Full & 16 & 20/20 & 1/20 & 24\\
Compact & 8 & 16/20 & 3/16 & 13\\
Compact & 16 & 20/20 & 3/20 & 15\\
\bottomrule
\end{tabular}
\end{table}

Naturally occurring recovery rates are selected by the model's preceding behavior. A model that makes fewer mistakes may leave a harder subset of errors in the recovery denominator. Equal total budgets do not imply equal budgets after the first error. A matched-error experiment now addresses this concern
(Section~\ref{sec:recovery}): it injects the same rejected call
after the same verified four-action prefix and fixes the remaining
budget. This compares post-error behavior under a standardized
disturbance. Its no-injection baseline follows model-generated
states and therefore does not estimate the causal effect of the
injected rejection; the experiment also covers one error type,
one adapter pair, and one task family.

\subsection{Untuned 8B chain reliability}

Error incidence and recovery further distinguish local correctness from completion. In untuned 8B chain runs, all twenty full-state episodes at each horizon contain invalid actions; only one H8 episode and no H16 episode eventually succeed. Under compact evaluation, four of nine error-containing H8 episodes and eight of twelve error-containing H16 episodes succeed.

\subsection{Untuned 27B chain reliability}
\label{app:27b}

We evaluated the untuned Qwen3.8-27B checkpoint on twenty paired full/compact tasks at each of H8 and H16. It uses the same task generator, controlled-pair interface, verifier, decoding settings, and $2H$ action cap as the matched untuned 8B protocol, but belongs to a different model generation and uses BF16 vLLM serving on an A100 cluster. Table~\ref{tab:27b} separates canonical oracle agreement, eventual completion, and success following invalid actions.

All twenty full-state H16 free runs contain an invalid call, although sixteen eventually succeed; only two compact-state runs contain an invalid call. At H8, eleven full-state episodes contain errors and ten of those succeed, versus one error-containing compact-state episode. Valid JSON is 100\% in every condition. At H16 in free runs, the legal-action rate is 68.13\% under full observations and 99.38\% under compact observations. Every recorded disagreement with the canonical target in the full-state conditions is also illegal, unlike a legal alternative order on the branching tasks; the aggregate error localization does not establish which binding operation caused the invalid call. Appendix~\ref{app:statistics} reports exploratory paired full/compact contrasts. The 27B checkpoint and smaller-model cases are complementary, not an isolated parameter-scaling experiment.

The horizon-matched H16 adapter achieves 20/20 free-run successes with no observed errors in either view (Table~\ref{tab:27b-learning}), but it is evaluated on different task seeds under an NF4 plus LoRA local inference path. The before/after values therefore illustrate different observed execution patterns, not a paired estimate of the effect of fine-tuning. Recovery of one or two error-containing episodes does not establish a stable recovery rate; the error distributions and remaining budgets differ across conditions.

\section{Training-Intervention Pilot Details}
\label{app:augmentation}

The four conditions are full-only, compact-only, a
full/compact mixture, and a full/random mixture. Each
mixture assigns 400 examples to each view, presenting each
source example once rather than duplicating it across
views. Artifact names, recipes, and target actions remain
unchanged. Random masking may hide a required input artifact's
contents; supplied identifiers and recipes still suffice
for selection and invocation in these tasks. Matching
examples and their order does not match input-token counts
or compute. This pilot includes no explicit recovery
supervision or relevance-matched causal control.

The pilot uses the first 800 action rows of the original
H16 full-state training manifest (3,200 source rows),
covering 50 complete workflows. The four conditions keep
the same gold prefixes and completions. Assignment to a
mixture's augmented half is a deterministic hash ordering
of task fingerprint and stage with seed 2026092001; both
mixtures share this assignment. Among the 400 random-view
examples, 134 use masking probability 0.25, 133 use 0.50,
and 133 use 0.75. Masks are generated once before training
and remain fixed. A masked artifact's contents are replaced by
\texttt{available}; the underlying environment is unchanged.
Unlike compact masking, random and block masks do not
consult a target action.

All four runs use Qwen3-1.7B with BF16 LoRA, rank 16, alpha
32, dropout 0.05, AdamW, learning rate $2\times10^{-4}$,
microbatch one, gradient accumulation eight, and training
and data seed 20260920. The maximum sequence length is
2,048 and packing is disabled. The longest tokenized
training examples are 1,351, 476, 1,340, and 1,340 tokens
for full, compact, full/compact, and full/random training,
respectively. Preparation checks verify complete target
retention without truncation. Runtime exposure records
verify 800 distinct rows presented once per run, with eight
microbatches per optimizer step and identical row order
across the four runs. This is action-level SFT on correct
prefixes, without on-policy or rejection-recovery examples.

Evaluation uses ten H16 development tasks with seeds
91000--91009, whose fingerprints are disjoint from training
and the supplied 80-row development files. Decoding is
greedy, with thinking disabled and at most 160 generated
tokens per call and a 32-call budget for free-running
rollouts. Endpoint evaluations cross four adapters,
three views, and two modes on these tasks (240 episodes).
Earlier checkpoints at steps 20, 40, 60, and 80 use full and
compact oracle-prefix probes (320 episodes); the untuned
baseline adds 20 oracle-prefix episodes. The completed 580
episodes repeatedly evaluate ten task instances, not 580
independent tasks. The block mask uses seed 2026092002 and
selects a contiguous block of $\lfloor |A_t|/2\rfloor$
names from the sorted available names, replacing the
corresponding artifact contents with \texttt{available}. The block location depends
on the task fingerprint and available names, so matched
states receive matched masks; free-run states can diverge
across policies. This block rule is not used during training;
all evaluation views retain artifact names and candidate
recipes.

\begin{table}[h]
\centering
\small
\caption{Exploratory paired success contrasts for the
full/random mixture versus each baseline at step 100.
Counts are discordant tasks: success only for full/random,
or only for the baseline. Effects are percentage points
over ten paired tasks. Exact two-sided paired tests are
unadjusted for six comparisons and condition on the
realized adapters.}
\label{tab:augmentationpaired}
\begin{tabular}{llrrrr}
\toprule
Baseline & Test view & Random only & Baseline only & Effect & $p$\\
\midrule
Full only & Full & 1 & 3 & $-20$ & .625\\
Full only & Compact & 3 & 2 & $+10$ & 1.000\\
Full only & Block & 1 & 3 & $-20$ & .625\\
Full/compact & Full & 6 & 1 & $+50$ & .125\\
Full/compact & Compact & 0 & 3 & $-30$ & .250\\
Full/compact & Block & 4 & 0 & $+40$ & .125\\
\bottomrule
\end{tabular}
\end{table}

For each contrast we pair by task fingerprint and apply
an exact two-sided binomial test to the discordant success
counts, equivalent to a paired sign-flip test for binary
outcomes. These small-sample comparisons do not establish
equivalence or a general ranking of training strategies.
The pilot adds one full-trained and one compact-trained
adapter under a new seed, but its evaluation set and
development role differ from the original crossover. We
report it separately rather than pooling task outcomes or
treating it as a completed multi-seed confirmation.

\section{ScienceWorld Protocol and Additional Results}
\label{app:scienceworld}

\paragraph{Data and split roles.}
The experiments use ScienceWorld 1.3.0 and five task
families: \texttt{boil}, \texttt{melt}, \texttt{freeze},
\texttt{chemistry-mix}, and \texttt{grow-plant}. The initial
fine-tuning dataset contains 39 trajectories and 1,170
action examples, with 13 DEV trajectories and 440 action
examples. Full and compact renderings share target actions.
These are source dataset counts, not an execution audit of
post-preprocessing rows or distinct consumed examples.
The reported configuration uses NF4 QLoRA, rank 16,
learning rate $2\times10^{-4}$, effective batch size eight,
training seeds 20260910--20260912, and checkpoints at steps
20, 40, 60, 80, and 100. The DEV set supports checkpoint
selection and is not an untouched generalization test.

The initial representation pilot evaluates ten official
test variations under three seeds and four train/test-view
cells, giving 120 rollouts. No official-test gold paths
are accessed. Its local DEV estimates and autonomous
official-test estimates involve different task sets.

The later natural-horizon study builds a gold-path catalog
from 120 official TRAIN and 60 official DEV variations.
Of these, 119 and 55 respectively reach score 100 within
100 reference actions. Its short-only training set contains
11 complete workflows and 218 action examples; checkpoint
selection uses four short workflows and 80 examples.
The selected short-only checkpoints are steps 100, 60,
and 100 for the three seeds, respectively, using the
reported prespecified DEV selection rule. All seeds are
retained. The mixed-horizon comparison reuses the original
full-view mixed-length adapters. Exact checkpoint identities
and selection details require the run manifests.

Natural-horizon bins are defined by reference-path length:
short has at most 24 actions, medium has 25--40, and long
has 41--100. The evaluation protocol freezes 30 workflows
unused by the initial train/DEV selection, with ten per bin;
the experiment summary identifies them as held out from
training. They come from official TRAIN/DEV, not official
TEST. Short evaluation workflows comprise four
\texttt{chemistry-mix}, two \texttt{freeze}, and four
\texttt{melt} tasks; medium workflows comprise three
\texttt{boil}, two \texttt{freeze}, three \texttt{grow-plant},
and two \texttt{melt} tasks; long workflows include two
from each family. Natural length therefore covaries with
task semantics. The bins use reference-path lengths, not
proven minimal solution lengths or the agent's actual calls.

The same 1,042 reference states are evaluated for each of
six model/seed configurations, yielding 6,252 local
predictions. The corresponding autonomous evaluation has
180 rollouts on the same 30 workflows. It is distinct from
the 120-episode official-test representation pilot. Each
rollout allows at most 100 calls; success means reaching
score 100, and mean final score is reported separately.

\begin{table}[h]
\centering
\small
\caption{ScienceWorld gold-prefix DEV agreement (\%),
averaged across three training seeds on the same 440 states.
F/C denotes full/compact. These are learning diagnostics on
checkpoint-selection data with mixed natural horizons.}
\label{tab:scienceworldcurves}
\begin{tabular}{rrrr}
\toprule
Optimizer step & F $\to$ F & C $\to$ C & C $\to$ F\\
\midrule
20 & 54.17 & 51.67 & 53.48\\
40 & 66.89 & 64.85 & 64.70\\
60 & 71.74 & 71.44 & 70.45\\
80 & 77.73 & 77.27 & 73.79\\
100 & 79.39 & 78.64 & 76.89\\
\bottomrule
\end{tabular}
\end{table}

The unadapted checkpoint has 155/440 correct full-view
predictions and 154/440 compact-view predictions. The
compact-trained model's compact advantage is 3.48
percentage points at step 80 and 1.74 points at step 100,
using unrounded aggregates. Accuracy improves under both evaluation views.

\begin{table}[h]
\centering
\small
\caption{ScienceWorld step-100 representation crossover.
DEV agreement is the mean over three training seeds on
440 reference states. Pilot completion is measured on ten
different official-test tasks per seed. Local and autonomous
columns here are not paired on the same tasks.}
\label{tab:scienceworldcross}
\begin{tabular}{llrr}
\toprule
Train view & Test view & DEV agreement (\%) & Pilot completion\\
\midrule
Full & Full & 79.39 & 7/30\\
Full & Compact & 75.08 & 0/30\\
Compact & Full & 76.89 & 8/30\\
Compact & Compact & 78.64 & 6/30\\
\bottomrule
\end{tabular}
\end{table}

\begin{table}[t]
\centering
\small
\caption{ScienceWorld natural-horizon evaluation. Each row
contains ten distinct workflows evaluated under three
training seeds (30 rollouts). Local agreement is measured
on reference states; completion and mean final score come
from paired autonomous runs with a 100-call budget. The
same workflows are used for both training conditions.}
\label{tab:scienceworldhorizon}
\begin{tabular}{llrrr}
\toprule
Training & Length & Local (\%) & Completion & Mean score\\
\midrule
Short-only & Short & 87.02 & 26/30 & 87.50\\
Short-only & Medium & 65.40 & 15/30 & 71.17\\
Short-only & Long & 65.64 & 5/30 & 45.87\\
Mixed-horizon & Short & 81.89 & 13/30 & 67.77\\
Mixed-horizon & Medium & 82.96 & 9/30 & 64.80\\
Mixed-horizon & Long & 80.67 & 5/30 & 44.63\\
\bottomrule
\end{tabular}
\end{table}

\paragraph{Paired local/autonomous diagnostics.}
The natural-horizon study evaluates each checkpoint and
workflow both from reference states and in autonomous
execution. Short-only short-bin successes are 9/10, 7/10,
and 10/10 by seed, versus long-bin successes of 2/10, 0/10,
and 3/10. The corresponding local-accuracy decreases are
23.24, 20.65, and 20.26 percentage points. These seed-wise
patterns describe this sample, not a randomized length effect.

For short-only training, workflows with local agreement
at least 80\% account for 27/37 successful rollouts, versus
19/53 below that threshold. For mixed-horizon training,
the respective counts are 17/56 and 10/34. These are
seed-by-workflow observations. The reported Spearman
associations with binary success are 0.489 and 0.128;
associations with final score are 0.542 and 0.225. We report
these descriptive diagnostics without treating repeated
workflows or correlated action predictions as independent
samples for significance testing.

Only 3/90 short-only and 2/90 mixed-horizon evaluations
have clean oracle paths; all of these correspond to
successful autonomous runs. Many other tasks also succeed.
Thus perfect agreement in the separate reference-state
probes is not necessary for observed autonomous success.
This does not show which autonomous trajectories deviated
from the reference path or recovered from errors; that
requires inspecting their own executed calls.

\begin{table}[h]
\centering
\small
\caption{Observed parser recovery in the separate
ScienceWorld official-test pilot. Each cell has ten tasks
evaluated under three seeds. Recovery counts successful
episodes among those with the exact parser-rejection string;
later score increase is a separate progress measure.}
\label{tab:scienceworldrecovery}
\begin{tabular}{lrrr}
\toprule
Train/test & Rejection episodes & Later success & Later score increase\\
\midrule
Full/Full & 14/30 & 1/14 & 4/14\\
Full/Compact & 12/30 & 0/12 & 2/12\\
Compact/Full & 18/30 & 4/18 & 12/18\\
Compact/Compact & 19/30 & 6/19 & 9/19\\
\midrule
Total & 63/120 & 11/63 & 27/63\\
\bottomrule
\end{tabular}
\end{table}

A parser rejection is counted only when feedback is exactly
\texttt{No known action matches that input.}
Across the 120 pilot episodes, 21 succeed and 1,510 calls
receive this feedback. Recovery
counts by training seed are 5/20, 5/22, and 1/21. Different
policies induce different error cases, times, and remaining
budgets, so these fractions do not identify a causal effect
of a representation on recovery. The fraction of calls
without this exact feedback is only a parser-based measure;
it does not certify semantic validity or task progress.

\paragraph{Evidence and reproducibility limits.}
ScienceWorld results are reported from our Q1--Q5 aggregate
run summary. Arithmetic checks confirm the reported totals,
but raw trajectories and executed training settings have
not been independently verified against the full run bundle.
The exact full/compact rendering rules and information
access still require documentation from the implementation;
Equation~\ref{eq:compact} defines the artifact intervention,
not the ScienceWorld views. A frozen 58-task official-test
extension, comprising $58\times3\times4=696$ planned rollouts,
has not been run.

\section{Evidence Provenance}
\label{app:provenance}

The controlled workflow experiments retain task fingerprints, split manifests,
protocol files, checkpoint identifiers, and per-action trajectories. The
original H16 crossover statistics were recomputed from paired trajectories;
branching actions were checked for legality and verified progress rather than
scored solely by canonical order. The later H16 seed pairs, 8B crossover,
matched-error evaluation, content-dependent choice study, and horizon--state
controls have separate frozen task sets and outcome records. Development
sweeps and superseded runs are retained in the project audit but are not
counted as independent replications.

\paragraph{27B training audit.}
Saved training arguments, trainer states, logs, source manifests, and
training-time bytecode establish the completed step-100 checkpoints and the
exposure reported in Table~\ref{tab:27b-exposure}. Some H8 sampled indices and
token lengths were reconstructed; exact launch arguments and active H16
compute-replica counts could not be recovered. This training audit does not
independently verify all held-out evaluation trajectories. Untuned and tuned
27B outcomes should therefore be interpreted within their separately
documented protocols.

\paragraph{ScienceWorld provenance.}
Section~\ref{sec:scienceworld} and Appendix~\ref{app:scienceworld} use our
ScienceWorld run summaries. They distinguish checkpoint-selection DEV, the
official-test representation pilot, and the paired natural-horizon study. The
source aggregates pass arithmetic checks, but the cluster renderer and raw
trajectories needed for independent trajectory-level verification have not
been consolidated. The planned larger official-test confirmation set
contributes no results to this paper.

\paragraph{Repository pilot.}
Exploratory cross-file Python repository runs did not establish a usable
external baseline: the completed V1 and V2 conditions each reached 0/5 exact
successes, and V3.1 reached 0/1 per condition. We exclude them from the main
claims because these failures do not distinguish model limitations from
interface calibration.

\end{document}